\documentclass[twoside,journey]{IEEEtran}
\usepackage{makecell}
\usepackage{hyperref}
\usepackage{array}
\usepackage{graphicx,amssymb,amsmath}
\usepackage{multicol}
\usepackage[noadjust]{cite}
\usepackage{setspace}
\usepackage{subfigure}
\usepackage{graphicx}
\usepackage{float}
\usepackage {url}
\usepackage{stfloats}
\usepackage{amsthm,pifont}
\usepackage{flushend}
\usepackage{cases,subeqnarray}
\usepackage{bm,multirow,bigstrut}
\usepackage{amsmath, amsthm, amssymb}
\usepackage{textcomp}
\usepackage{latexsym,bm}
\usepackage{booktabs}
\usepackage{xcolor}
\usepackage{mathtools}
\usepackage{dsfont}
\usepackage{extarrows}
\usepackage{epsfig}
\usepackage{epsfig}
\usepackage{epstopdf}
\usepackage[noend]{algpseudocode}
\usepackage{algorithmicx,algorithm}
\usepackage{makecell}
\usepackage{colortbl}
\usepackage{booktabs}
\usepackage{graphicx} 
\usepackage{array}    
\usepackage{hyperref} 
\usepackage{booktabs}
\usepackage{enumitem}
\usepackage{diagbox}
\usepackage{tikz}
\usetikzlibrary{trees,shadows.blur}
\usepackage{forest}
\theoremstyle{plain}

\theoremstyle{plain}

\usepackage{amsmath}

\newcommand{\ignore}[1]{{{\color{yellow} }}}
\definecolor{blue-green}{rgb}{0.0, 0.87, 0.87}
\IEEEoverridecommandlockouts
\begin{document}

\title{{\huge Edge General Intelligence Through World Models and Agentic AI: Fundamentals, Solutions, and Challenges}}

\author{Changyuan Zhao, Guangyuan Liu, Ruichen Zhang, Yinqiu~Liu, Jiacheng Wang, Jiawen Kang,\\Dusit Niyato,~\IEEEmembership{Fellow,~IEEE}, Zan Li,~\IEEEmembership{Fellow,~IEEE}, Xuemin (Sherman) Shen,~\IEEEmembership{Fellow,~IEEE}, \\Zhu Han,~\IEEEmembership{Fellow,~IEEE}, Sumei Sun,~\IEEEmembership{Fellow,~IEEE}, Chau Yuen,~\IEEEmembership{Fellow,~IEEE}, \\and Dong In Kim,~\IEEEmembership{Life Fellow,~IEEE}

\thanks{C. Zhao is with the College of Computing and Data Science, Nanyang Technological University, Singapore, and CNRS@CREATE, 1 Create Way, 08-01 Create Tower, Singapore 138602 (e-mail: zhao0441@e.ntu.edu.sg).
}
\thanks{G. Liu, R. Zhang, Y. Liu, J. Wang, D. Niyato, and C. Yuen are with the College of Computing and Data Science, Nanyang Technological University, Singapore (e-mail: liug0022@e.ntu.edu.sg; ruichen.zhang@ntu.edu.sg; yinqiu001@e.ntu.edu.sg; jiacheng.wang@ntu.edu.sg; dniyato@ntu.edu.sg; chau.yuen@ntu.edu.sg).}
\thanks{J. Kang is with the School of Automation, Guangdong University of Technology, China. (e-mail: kavinkang@gdut.edu.cn).
}
\thanks{Z. Li is with the State Key
Laboratory of Integrated Services Networks, Xidian University, Xian 710071,
China (e-mail: zanli@xidian.edu.cn).}
\thanks{X. Shen is with the Department of Electrical and Computer Engineering, University of Waterloo, Canada (e-mail: sshen@uwaterloo.ca).
}
\thanks{Z. Han is with the Department of Electrical and Computer Engineering at the University of Houston, Houston, TX 77004 USA, and also with the Department of Computer Science and Engineering, Kyung Hee University, Seoul, South Korea, 446-701 (e-mail: zhan2@uh.edu).}
\thanks{S. Sun is with the Institute for Infocomm Research, Agency for Science,
Technology and Research, Singapore (e-mail: sunsm@i2r.a-star.edu.sg).}
\thanks{D. I. Kim is with the Department of Electrical and Computer
Engineering, Sungkyunkwan University, Suwon 16419, South Korea (e-mail: dongin@skku.edu).}
}

\maketitle
\vspace{-1cm}

\begin{abstract}

Edge General Intelligence (EGI) represents a transformative evolution of edge computing, where distributed agents possess the capability to perceive, reason, and act autonomously across diverse, dynamic environments. Central to this vision are world models, which act as proactive internal simulators that not only predict but also actively imagine future trajectories, reason under uncertainty, and plan multi-step actions with foresight. This proactive nature allows agents to anticipate potential outcomes and optimize decisions ahead of real-world interactions. While prior works in robotics and gaming have showcased the potential of world models, their integration into the wireless edge for EGI remains underexplored. This survey bridges this gap by offering a comprehensive analysis of how world models can empower agentic artificial intelligence (AI) systems at the edge. We first examine the architectural foundations of world models, including latent representation learning, dynamics modeling, and imagination-based planning. Building on these core capabilities, we illustrate their proactive applications across EGI scenarios such as vehicular networks, unmanned aerial vehicle (UAV) networks, the Internet of Things (IoT) systems, and network functions virtualization, thereby highlighting how they can enhance optimization under latency, energy, and privacy constraints. We then explore their synergy with foundation models and digital twins, positioning world models as the cognitive backbone of EGI. Finally, we highlight open challenges, such as safety guarantees, efficient training, and constrained deployment, and outline future research directions. This survey provides both a conceptual foundation and a practical roadmap for realizing the next generation of intelligent, autonomous edge systems.

\end{abstract}
\IEEEpeerreviewmaketitle

\section{Introduction}\label{intro}



\subsection{Background}

Modern communication networks are evolving toward next-generation paradigms such as 5G, 6G, and beyond, enabling the interconnection of a massive number of devices at the network edge.
According to Statista, the global number of the Internet of Things (IoT) devices is projected to increase from 19.8 billion in 2025 to over 40.6 billion by 2034\footnote{https://www.statista.com/statistics/1183457/iot-connected-devices-worldwide/
}.
This explosive growth of devices at the edge has catalyzed a fundamental shift from centralized cloud computing to distributed edge computing~\cite{abkenar2022survey, kong2022edge, duan2022distributed}. Edge computing and edge intelligence have emerged as a compelling response to the limitations of cloud-only approaches, including high latency, bandwidth costs, privacy concerns, and scalability bottlenecks~\cite{barbuto2023disclosing, xu2021edge, zhou2019edge, mao2023security}.

However, current edge intelligence solutions remain largely task-specific. They excel at isolated functions, such as object detection, but lack the cognitive depth required for autonomous decision-making in dynamic, multi-modal environments~\cite{deng2020edge}. To bridge this gap, researchers have proposed the concept of Edge General Intelligence (EGI), 
 which aims to provide cloud-like versatility at the edge while retaining its low-latency and privacy-preserving advantages~\cite{luo2025toward}. 
 EGI is envisioned as a holistic architecture that integrates perception, prediction, and control across heterogeneous edge nodes, and can flexibly incorporate cloud resources through centralized, hybrid, or fully decentralized deployments~\cite{chen2025towards}.
To realize this vision of EGI, recent advancements have focused on the development of artificial intelligence (AI) agents and, more broadly, agentic AI. These systems are designed to perceive, reason, plan, and adapt autonomously in complex and dynamic environments~\cite{acharya2025agentic}.
Generally, agentic AI is based on large language models (LLMs) or foundation models with tool use, planning, and memory, offering a promising foundation~\cite{sapkota2025ai,he2025road}.

However, recent studies have demonstrated that language-based reasoning has fundamental limitations in capturing high-dimensional, multi-modal environmental dynamics~\cite{saad2025artificial,yang2025thinking, lecun2022path}.
As Yann LeCun illustrates, consider the task of understanding a cube rolling on a table. Humans can intuitively and accurately predict its motion, yet describing that same process in natural language is inherently imprecise and incomplete\footnote{https://ai.meta.com/vjepa/}. Feifel Li’s World Labs similarly champions spatial intelligence that grounds LLM reasoning in a 3D physical context\footnote{https://www.worldlabs.ai/}.
Popular science fiction films such as \textit{The Matrix} and \textit{Inception} dramatize this challenge by presenting agents operating within richly simulated environments, where success depends on their ability to model and predict the consequences of actions within a physically coherent virtual world.
This idea is now gaining significant attention in AI research: \textit{how can AI, like humans, learn the fundamental laws governing our world and make informed predictions?}
This observation motivates the integration of world models, \textit{an internal, predictive simulator of how the environment will evolve under different actions}. 
A typical world model is instantiated as a deep neural architecture comprising \textit{(i)} an encoder that compresses raw sensor streams into latent codes, \textit{(ii)} a dynamics module that predicts latent transitions conditioned on actions, and \textit{(iii)} a decoder that reconstructs observations or task-specific signals. 
These neural architectures are trained end-to-end using techniques such as variational inference, contrastive learning, and reconstruction losses to learn predictive models of environment dynamics~\cite{ha2018world, hafner2023mastering}.

\begin{figure}[t]
    \centering
    \includegraphics[width= 0.95\linewidth]{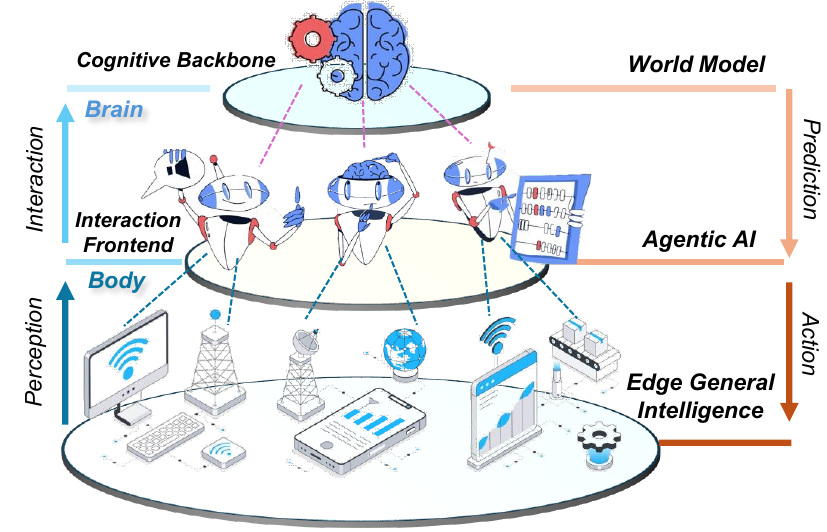}
    \caption{The hierarchical architecture of EGI driven by world models. The cognitive backbone (world model) proactively predicts future dynamics, while the interaction frontend (agentic AI) perceives the environment and executes actions. This perception–prediction–action loop enables proactive reasoning and adaptive decision-making in dynamic edge scenarios.
    }
    \label{fig:fig_one}
\end{figure}

Unlike language-based reasoning alone, world models provide temporal foresight, spatial reasoning, and counterfactual thinking~\cite{ding2024understanding}. 
Therefore, world models serve as \textit{the cognitive backbone} or ``brain" of the AI agent, while the agentic AI framework functions as \textit{the interaction frontend} or ``body" to sense, act, and interact with the real world~\cite{garrido2024learning}.
Leveraging this paradigm, agentic AI continually interacts with the backend world model, querying it for predictions, simulations, and strategic planning to inform its decisions in complex environments (Fig. \ref{fig:fig_one}).
Notably, Meta's V-JEPA v2 advances this paradigm by proposing a ``blueprint for general intelligence” that leverages a world model to perceive physical reality, anticipate future outcomes, and formulate efficient action strategies~\cite{assran2025v}.
In summary, world models enhance agentic AI with temporal prediction, spatial reasoning, and counterfactual thinking capabilities that language alone cannot provide. They enable AI agents to think beyond words and act with foresight in complex environments.

\subsection{Motivation}


The vision of EGI demands foresight, sample efficiency, and safety that today’s task-specific edge intelligence pipelines cannot adequately address~\cite{he2025road}. These requirements become even more pronounced in wireless network management, where heterogeneous edge nodes, including on-device, edge-server, and cloud components, must collaborate under stringent latency and reliability constraints~\cite{chen2025towards}. 
They are inherently high-dimensional, partially observable, and non-stationary~\cite{liu2024survey}. 
Channel coherence times may drop below 1 ms, and user mobility quickly renders prior measurements obsolete. Moreover, safety-critical applications, such as vehicular platooning, cannot tolerate long exploration phases that disrupt service~\cite{xu2025fully,xu2022channel,10016213,yang2024navigating}.
More broadly, edge scenarios exacerbate the limitations of conventional AI systems. These systems struggle with non-stationary dynamics, severely constrained opportunities for real-world interaction, and extreme spatiotemporal variability~\cite{zhao2025world}. In such settings, collecting sufficient on-device data is often infeasible due to safety, latency, or energy constraints. Critically, traditional AI approaches are unable to generalize to out-of-distribution conditions not encountered during training~\cite{zhao2024generative2}. Without the capacity to simulate and reason over plausible future trajectories, these systems falter in environments where adaptability and anticipatory decision-making are essential.

World models offer a principled solution to these constraints. By learning an internal simulator of latent physical dynamics, an intelligent agent can imagine thousands of counterfactual scenarios offline, reason about long-horizon trade-offs, and deploy only a minimal number of on-device interactions to refine its policy~\cite{hao2023reasoning}. 
Unlike conventional predictive models that passively respond to incoming data, a world model is inherently \textit{proactive}. 
This imagination-driven approach fundamentally transforms the expensive ``trial-and-error" cycle of model-free reinforcement learning into a low-cost ``imagine-and-action" paradigm, better suited to the computational heterogeneity and environmental constraints of complex edge scenarios~\cite{gao2024vista}.
Recent work on Wireless Dreamer demonstrates how coupling latent-dynamics models with imagination-based planning can guide unmanned aerial vehicle (UAV) trajectory optimization in low-altitude networks, achieving faster convergence and higher throughput than model-free approaches in weather-aware communications~\cite{zhao2025world}. This paradigm can be further extended to modeling long-horizon 3D interference patterns in UAV networks, offering a potential foundation for interference-aware trajectory and resource planning~\cite{sharma2019random}. Similarly, in millimeter-wave networks, world model-based schedulers have reduced packet-completeness-aware age of information, even under severe link blockages and sub-millisecond coherence constraints~\cite{wang2025world}. 
These early successes demonstrate that world models can successfully translate their proven benefits from robotics and gaming domains to mission-critical wireless optimization. 
Finally, world models encode environmental features into a compact latent space and perform dynamic inference within this latent domain. This design allows agents to capture essential temporal and spatial structures while avoiding the computational burden of operating in the raw observation space~\cite{ha2018world}. As a result, world models introduce the capability of ``trial-and-error" learning while preserving efficiency, enabling scalable and adaptive decision-making for wireless edge scenarios.

In summary, leveraging the imagination capabilities of world models offers the following potential advantages:
\begin{itemize}
    \item \textbf{Foresight and Efficiency:} Offline imagination reduces real-world interactions and accelerates policy learning.

\item \textbf{Robustness to Dynamics:} Learned latent models enable adaptation to non-stationary wireless conditions.

\item \textbf{Safe Decision-Making:} Supports anticipatory actions with minimal risk, ideal for mission-critical applications.

\end{itemize}

Despite the parallel evolution of EGI architectures and world model methodologies, these two research streams remain largely disconnected. Existing surveys either consider edge computing frameworks without dynamics-aware modeling approaches or review world model advances primarily from computer vision or robotics perspectives~\cite{luo2025toward, feng2025survey, guan2024world}. To our knowledge, no comprehensive analysis exists that systematically explores how core world model principles can be effectively integrated into the optimization stack of EGI.

\textit{This survey aims to fill that gap by offering a unified perspective.} We provide both conceptual and practical guidance on embedding world models into edge optimization tasks across diverse scenarios.
This survey paper can benefit ongoing and emerging research studies in EGI in the following ways:
\begin{itemize}
    \item \textbf{World models as the Cognitive Core:}  
    Position world models as the ``brain” and agentic AI as the ``body,” enabling foresight, reasoning, and planning in edge systems.
    
    \item \textbf{Grasp the Core Architecture:}  
    Understand key modules of world models and how they interact to support long-horizon decision-making.
    
    \item \textbf{Compare Model Variants:}  
    Examine representative frameworks, highlighting differences in design, planning capability, and edge suitability.
    
    \item \textbf{Adapt to Edge Wireless Scenarios:}  
    Show how existing world models can be tailored to edge tasks under partial observability and strict constraints.
    
    \item \textbf{Challenges and Future Directions:}  
    Discuss open issues, such as computational bottlenecks, real-time planning constraints, and robustness.
\end{itemize}


\begin{table*}[htp]
\scriptsize
\centering
\caption{SUMMARY OF RELATED SURVEYS}
\label{tab:related}
\begin{tabular}{m{0.12\textwidth}<{\centering}||m{0.08\textwidth}<{\centering}|m{0.14\textwidth}<{\centering}|m{0.55\textwidth}}
    \hline
    \textbf{Scope} & \textbf{Reference}  & \textbf{Emphasis} & \multicolumn{1}{c}{\textbf{Overview}}  \\
    \hline
    \multirow{3}{0.12\textwidth}[-5pt]{\centering Edge Intelligence} 
    & \cite{xu2020edge} & Edge intelligence Fundamentals & A survey introducing a four-pillar framework for deploying AI efficiently at the network edge\\
    \cline{2-2} \cline{3-4}
    & \cite{barbuto2023disclosing} & Meta-survey on edge intelligence & 
    A review of edge intelligence, tracing its past developments, current state, and future directions, as well as its interconnections with the IoT and cloud computing systems\\
    \cline{2-2} \cline{3-4}
    & \cite{luo2025toward} & EGI via Multi-LLMs & 
    An overview of architectural designs and deployment strategies tailored for multi-LLM systems operating in edge computing environments\\
    \hline
    \multirow{4}{0.12\textwidth}[-15pt]{\centering World Model} 
    & \cite{guan2024world} & Autonomous Driving &  
    An initial review of world models in autonomous driving, spanning
their theoretical underpinnings, practical applications, and ongoing research efforts
\\
    \cline{2-2} \cline{3-4}
    & \cite{feng2025survey} & Autonomous Driving & 
    A technical roadmap for harnessing the transformative potential of world models in advancing safe and reliable autonomous driving solutions\\
    \cline{2-2} \cline{3-4}
    & \cite{zhu2024sora} & General World Models & A review of General World Modeling to video generation, autonomous driving, and intelligent agents \\
    \cline{2-2} \cline{3-4}
    & \cite{ding2024understanding} & World Models Paradigm Analysis & A review of world models categorized by their functions in representation learning and future prediction \\
    \hline
\end{tabular}
\end{table*}

\subsection{Related Surveys and Contributions}

\subsubsection{Edge Intelligence}

Recent literature has witnessed a significant surge in the exploration of edge intelligence and EGI (Table \ref{tab:related}). 
The work in \cite{xu2020edge} offers an early comprehensive survey on edge intelligence, identifying four core pillars: caching, training, inference, and offloading. It analyzes their architectural designs, optimization objectives, and privacy challenges in deploying AI at the edge.
\cite{barbuto2023disclosing} further extends this view with a systematic meta-survey, highlighting edge intelligence cross-domain relevance to IoT, cloud–edge continuum, and real-time intelligence. The study classifies research trends, enabling technologies, and emphasizes the integration of intelligence into resource-constrained edge environments.
Meanwhile, 
Finally, \cite{luo2025toward} explores the use of multi-LLMs to realize ubiquitous EGI. Their survey details how multiple specialized LLMs, when dynamically orchestrated, can overcome the limitations of single models in heterogeneous edge environments. 

\subsubsection{World Model}

The survey in \cite{guan2024world} focuses on autonomous driving, identifying key components of world models, including perception, memory, prediction, and control, and emphasizing their role in long-horizon planning and uncertainty-aware reasoning.
Similarly, \cite{feng2025survey} introduces a three-level taxonomy, i.e., scene generation, behavior planning, and planning-prediction interaction, while addressing challenges such as multimodal fusion, long-tail generalization, and real-time control.
Furthermore, \cite{zhu2024sora} broadens the scope to general world models, framing video generation, autonomous agents, and embodied AI under a unified generative world modeling paradigm, with Sora and DriveDreamer highlighted as milestones.
Lastly, \cite{ding2024understanding} classifies world models into two paradigms, including internal representation and future prediction. It connects classical latent models with recent LLM-based simulators, aiming to bridge perception, reasoning, and planning.

Distinct from existing surveys and tutorials, our survey distinguishes itself by specifically focusing on the integration of world models in agentic AI within Edge General Intelligence architectures. 
Unlike previous works, which either broadly address edge intelligence frameworks without dynamics-aware modeling approaches or review world model advances primarily from other perspectives, this survey offers a unique perspective by marrying the cognitive capabilities of world models with the operational requirements of EGI systems. 
The key contributions of this paper are summarized as follows:
\begin{itemize}
    \item We present the first comprehensive survey that systematically links world models to EGI of future edge systems, thereby closing the gap between dynamics-aware modelling and edge-scale autonomy.
    \item We provide a foundational overview of world model methodologies. We highlight how these models enable agents to capture latent dynamics, predict multi-step futures, and support decision-making under uncertainty.
    \item We examine the applicability of world models across diverse EGI wireless scenarios, demonstrating how their imagination capabilities can enhance wireless optimization tasks, such as power allocation in vehicular networks and link scheduling in UAV networks.
    \item We identify key open challenges and opportunities in deploying world models for EGI. These insights establish a roadmap for future research in the convergence of EGI and next-generation wireless systems.
\end{itemize}

\begin{figure}[t]
    \centering
    \includegraphics[width= 0.85\linewidth]{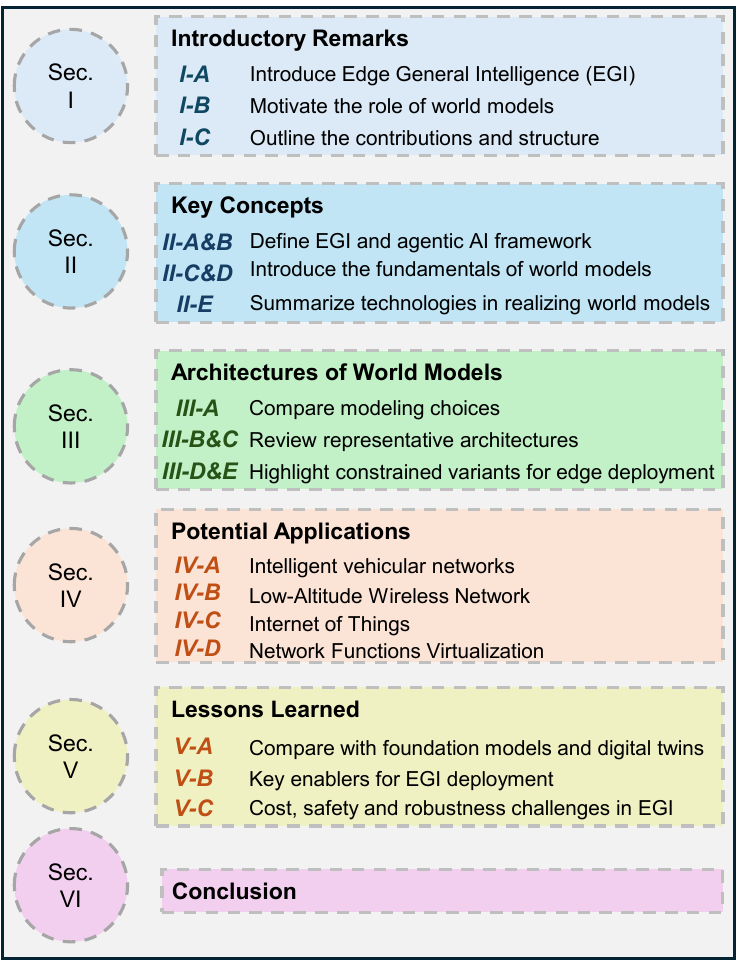}
    \caption{This figure outlines the structure, with each box summarizing a section’s key themes and contributions.
    }
    \label{fig:structure}
\end{figure}

The structure of this survey is outlined in Fig. \ref{fig:structure}.

\section{Overview of Edge General Intelligence and World Models}

\begin{figure}[t]
    \centering
    \includegraphics[width= 0.95\linewidth]{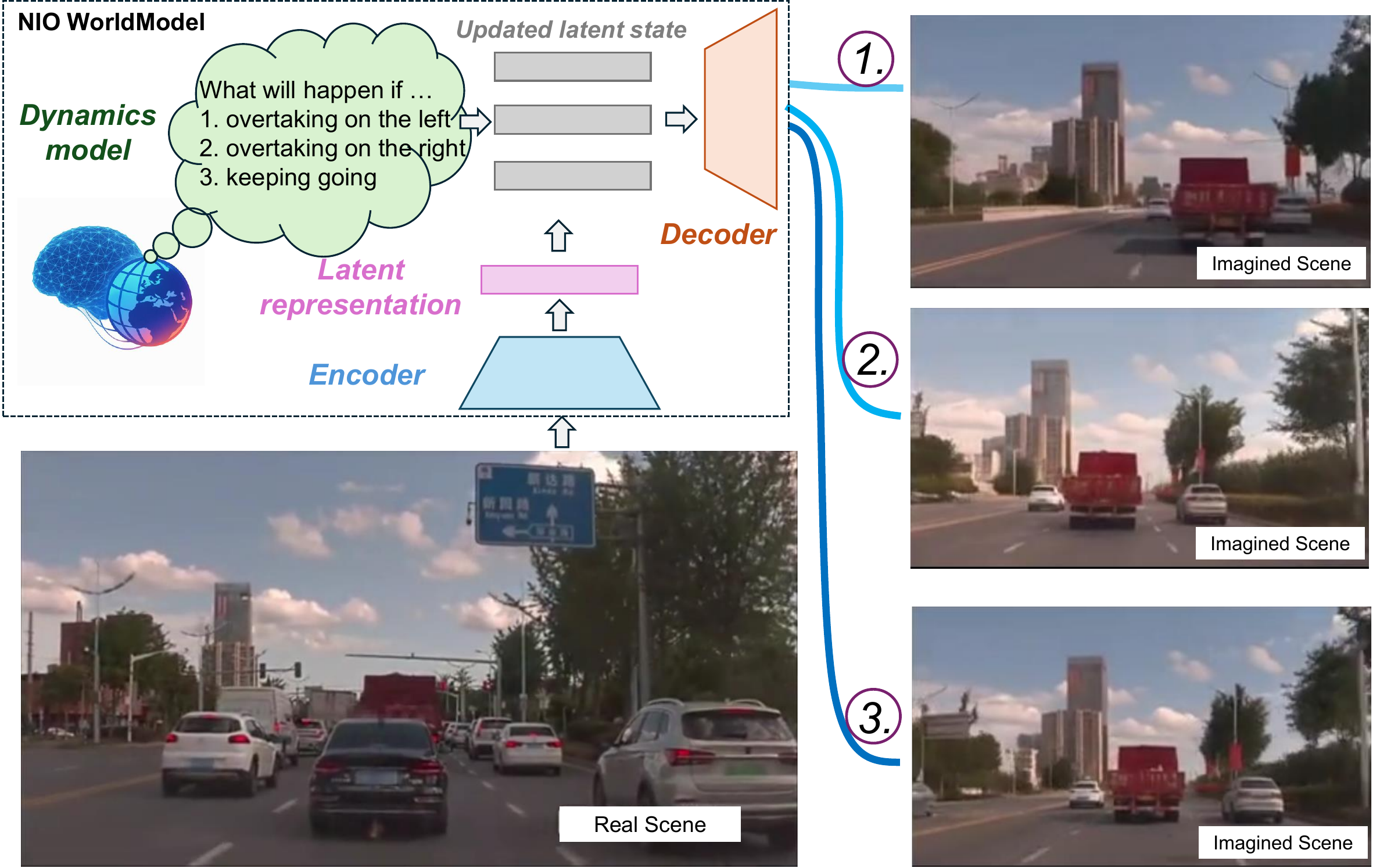}
    \caption{A potential architecture of NWM inspired by \cite{ha2018world}, including encoder, dynamic modeling, and decoder components. 
    The encoder compresses the real scene into a latent representation, which the dynamics model uses to imagine alternative futures. Steps 1–3 illustrate predicted outcomes for different actions.
    }
    \label{fig:NWM}
\end{figure}

\begin{figure*}[t]
    \centering
    \includegraphics[width= 0.95\linewidth]{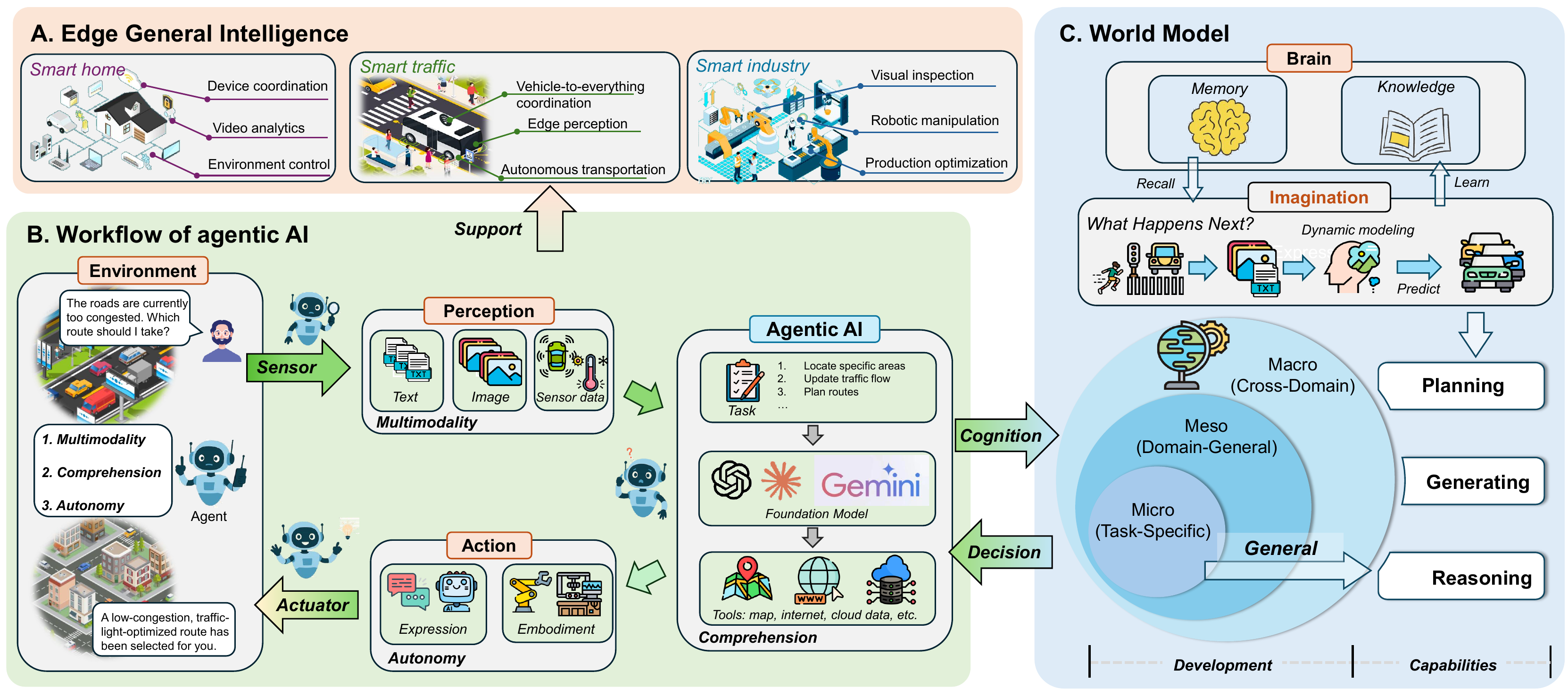}
    \caption{Illustration of world model, agentic AI, and EGI composition and development timeline.
    \textit{Part A:} EGI applications in smart homes, traffic, and industry with distributed intelligence capabilities.
   \textit{Part B:} Agentic AI system showing perception-cognition-action loop with multimodal processing.
   \textit{Part C:} World model framework with hierarchical cognitive architecture from task-specific to cross-domain capabilities.
    }
    \label{fig:overreview}
\end{figure*}

\subsection{Edge General Intelligence}
\label{sec:egi}

EGI refers to the next evolution of edge computing and AI, where edge devices exhibit human-like general cognitive abilities across a broad range of tasks. 
In contrast to traditional edge intelligence, which has mostly been narrow, specialized models for single tasks, EGI aspires to general intelligence at the edge \cite{chen2025towards, he2025road, luo2025toward}. An EGI system would possess high-level cognitive capabilities, including comprehension, multimodal perception, and autonomy, analogous to human intelligence \cite{chen2025towards}. 
In practical terms, EGI means that heterogeneous edge nodes, such as roadside units in vehicular networks and 5G base stations in dense urban deployments, can collaboratively integrate perception, prediction, and control to understand complex instructions, learn and accumulate knowledge over time, and adapt to dynamic environments and diverse tasks while flexibly leveraging cloud resources when needed~\cite{he2025road, qu2024digital}.

Recent advancements in LLMs and foundation models have significantly improved the practicality of centralized EGI systems, with major AI service providers providing ready-to-use application programming interfaces (APIs) that simplify the integration of these models into existing edge systems~\cite{chen2025towards}.
Leveraging the highly effective generalization, LLMs and foundation models are increasingly used as autonomous agents that can understand context, plan, take actions through tools, and reflect on their progress, supporting tasks such as literature search, code generation, and project management.
For example, the Auto-GPT system that chains web search, execution, and self-feedback to deliver business analyses and software prototypes \cite{yang2023auto}.
This has led to a surge in interest and innovation in EGI, which addresses key industrial needs such as agile connectivity, real-time services, and data optimization~\cite{friha2024llm}.
EGI has the potential to transform various sectors through its applications, including real-time video analytics, cognitive assistance, precision agriculture, smart cities, and industrial Internet of Things (IoT) systems, as shown in Fig. \ref{fig:overreview} \textit{Part A}. Companies such as Google, Microsoft, IBM, and Intel have been at the forefront of demonstrating how edge computing can enhance AI applications and facilitate the ``last mile" of AI deployment\footnote{https://viso.ai/edge-ai/edge-intelligence-deep-learning-with-edge-computing/}.

While EGI is promising, realizing it in practice faces significant technical and practical challenges:
\begin{itemize}
    \item \textbf{Resource Constraints:} Edge devices have limited computation, memory, and power, yet EGI’s general-purpose models are massive. Even quantized and optimized versions of large models can require on the order of gigabytes of memory and considerable FLOPs, which is orders of magnitude beyond typical embedded capacities~\cite{chen2025towards, wang2025energy}. 
    For instance, after aggressive optimization, models such as Llama 3.2 1B may still require more than 500MB of RAM when quantized to 4-bit precision, pushing the limits of mobile hardware \cite{touvron2023llama}. 
    
     \item \textbf{Extreme Environments and Reliability:} Edge deployments in remote or harsh environments, such as offshore platforms, disaster zones, and battlefields, face unreliable connectivity, limited power, and physical risks \cite{onsu2024leveraging}. In such edge extremes, EGI could be transformative, but the environmental conditions make it hard for agents to collaborate or get cloud help. Intermittent networks or power constraints mean an edge intelligent agent must operate gracefully offline, handling long periods of autonomy \cite{wei2025trustworthy}. Robustness against failures is critical when human oversight is minimal \cite{zhao2025generative}.
\end{itemize}

\subsection{Agentic AI}

To achieve EGI, agentic AI refers to AI systems that demonstrate autonomous, goal-directed behavior~\cite{jiang2025large}. Namely, AI agents that operate within various environments such as games, the web, or robotics labs \cite{acharya2025agentic}.
As depicted in Fig. \ref{fig:overreview} \textit{Part B}, agentic AI generally utilize a perception-cognition-action loop, enabling autonomous, goal-directed behavior in edge environments \cite{sapkota2025ai}.
This sophisticated framework mirrors biological cognitive processes, creating intelligent agents capable of environmental perception, complex reasoning, and adaptive action selection without human intervention\footnote{https://markovate.com/blog/agentic-ai-architecture/}. 

Agentic AI systems generally comprise three core modules working in concert to achieve perception-cognition-action loop.
\begin{itemize}
    \item \textbf{Perception Module:} Handles environmental awareness through multimodal integration of vision, audio, text, and sensor data. Advanced computer vision and natural language processing techniques process multi-source sensory inputs using raw data combination, decision-level integration, and hybrid approaches that balance computational efficiency with environmental representation fidelity \cite{he2025road}.

    \item \textbf{Cognition Module:} Serves as the central reasoning engine, typically implemented through LLMs that orchestrate decision-making processes. Goal representation systems manage dynamic objective setting and priority allocation, while planning modules employ Monte Carlo Tree Search and A* algorithms for optimal action sequence generation\footnote{https://blogs.nvidia.com/blog/what-is-agentic-ai/}. Recent architectures such as MemGPT enable extended context retention, and Retrieval-Augmented Generation (RAG) provides dynamic knowledge integration from external sources \cite{packer2023memgpt}.

    \item \textbf{Action Module:} Translates cognitive decisions into executable actions through actuators, APIs, and system interfaces. Tool integration capabilities enable dynamic access to external databases and specialized software, while feedback processing systems observe outcomes for continuous optimization. Adaptive control mechanisms adjust strategies based on environmental responses and performance metrics \cite{gridach2025agentic}.

\end{itemize}

The cognition module represents the most critical component for enabling planning and foresight in agentic AI. 
World models, which simulate potential action consequences before execution, empower agents to behave strategically and safely rather than merely react to stimuli \cite{ha2018world}. This proactive capability becomes increasingly essential as tasks grow in complexity and temporal scope, enabling agents to achieve high-level, long-term goals, as shown in Fig. \ref{fig:overreview} \textit{Part C}. 
Therefore, world models can serve as the cognitive backbone, while agentic AI functions as the interaction frontend, forming a closed loop of perception, prediction, and decision-making. 
The following sections explore how world models and optimization techniques enhance the cognition module to realize EGI.

\subsection{Definition and Core Components of World Models}

A world model generally refers to \textit{an internal, predictive simulator of how the environment will evolve under different actions} \cite{ha2018world}. 
By learning an internal representation of the environment’s physics, spatial dynamics, and causal structure, the world model enables an agent to simulate future states and outcomes, supporting planning and decision-making without needing to physically experience every outcome~\cite{ding2024understanding}.

Taking the NIO WorldModel\footnote{https://www.nio.cn/smart-technology/20241120002} (NWM) as an example, 
NWM reconstructs input of raw sensor information, automatically learning knowledge and physical laws from it, through autoregression.
Then, NWM uses the learned knowledge to predict and generate new possible future situations based on current sensor information.
Specifically, NWM can deduce 216 possible trajectories and find the optimal path within 100 milliseconds.
The images generated by NWM and the interactions between objects align with the fundamental principles of the physical world. NWM's focus is primarily on ``key points," such as an approaching vehicle that is about to merge, rather than on details such as the text or images on billboards. Therefore, it can generate a 120-second imaginary video based on a prompt input of a 3-second video.

Most world models are implemented as neural architectures with three core components that achieve the perception-prediction cycle \cite{ding2024understanding}:
\begin{itemize}
    \item \textbf{Encoder:} The encoder network compresses high-dimensional observations, such as images, sensor data, etc., into a compact latent state.
    For example, in the first world models system \cite{ha2018world}, a variational autoencoder (VAE) compresses 2D game frames into a 32-dimensional latent vector for strategic planning in games. 
    As shown in Fig. \ref{fig:NWM}, with the encoder network, NWM
    can transform the real scenes into a latent representation, discarding irrelevant details and retaining features crucial for prediction. 

    \item \textbf{Dynamics Model:} The dynamics module predicts how the latent state evolves over time, typically based on the agent’s actions. 
    In essence, this component learns the state transition function of the environment.
    For instance, recurrent mixture-density networks \cite{ha2018world} and recurrent state-space models (RSSMs) \cite{hafner2019learning} are employed to predict state transitions by leveraging the temporal learning capabilities of recurrent neural networks (RNNs).
    With the dynamics model (Fig. \ref{fig:NWM}), NWM updates the latent state from one time step to the next based on the car's actions, such as overtaking or keeping going straight.

    \item \textbf{Decoder:} The decoder network maps the latent state back to the observable output, reconstructing predicted observations or other quantities of interest.
    It closes the loop by translating the model’s latent imagination into the actual prediction of what the agent would observe next. 
    For example, the authors in \cite{ha2018world} leverage the VAE decoder to reconstruct each frame to visualize the quality of the predicted information.
    As depicted in Fig. \ref{fig:NWM}, the decoder generates the imagined scenes based on the car's actions from the updated latent states for the best driving strategy.
    
\end{itemize}

\subsection{Workflow of World Models}

The world model is an internal model learned by agents of how the environment changes in response to actions, and then uses this model to simulate trajectories in its “mind” without actual physical execution. 
Fig. \ref{fig:NWM} illustrates the typical workflow of using a world model for prediction and imagination \cite{hafner2019dream}. First, the agent interacts with the environment and collects a dataset of observations, actions, and rewards. Next, it trains the world model on this data by predicting the next state based on the states and actions.
Once trained, the world model functions as a proxy environment. The agent can input a candidate action to the model and predict what would happen. Crucially, unlike an open-loop forecast, the imagination can be goal-directed, where the agent can try many action sequences in the model and select the one with the highest predicted reward \cite{guan2024world}. 
In effect, the world model endows the agent with a form of counterfactual reasoning: “What if I do X? My model predicts Y, which is good/bad.”

Formally, consider an agent operating in a Markov Decision Process (MDP) with state $s_t$, action $a_t$, reward $r_t$, and dynamics $s_{t+1}=T(s_t,a_t)$. Since $T$ is unknown and observations $o_t$ may be high-dimensional, the world model learns an approximate transition $\hat{T}_\theta$ and reward function $\hat{R}_\theta$ so that $s_{t+1}\approx \hat{T}_\theta(s_t,a_t)$ and $\hat{R}_\theta(s_t,a_t)\approx r_t$. In practice, it maintains a latent state $h_t$ summarizing past observations and learns encoders, decoders$h_t = f_{\text{enc}}(h_{t-1}, o_t, a_{t-1})$ and $\hat{o}_t = f_{\text{dec}}(h_t)$ alongside a transition function $h_{t+1}=f_{\text{dyn}}(h_t,a_t)$ \cite{hafner2019dream}.
Training minimizes prediction error $||s_{t+1}-\hat{T}_\theta||$ or maximizes likelihood $P_\theta(o_{t+1},r_t|h_t,a_t)$ over experience data.
Once learned, the model can roll out trajectories internally. From $h_t$, it predicts $h_{t+1}$, $\hat{o}_{t+1}$, and $\hat{r}_t$ for action $a_t$, repeating for future steps to generate
\[
\hat{\tau}=(h_t,a_t,\hat{r}_t;h_{t+1},a_{t+1},\hat{r}_{t+1};\dots).
\]
The agent evaluates these imagined trajectories to optimize a policy $\pi(a|s)$ maximizing the expected cumulative reward under the model,
\[
\mathbb{E}_{\hat{T}_\theta}\Big[\sum_{k=0}^{H-1} r_{t+k}\Big].
\]
Planning can use direct search over action sequences, model-predictive control \cite{rawlings2000tutorial}, or gradient-based policy optimization using imagined rollouts \cite{matsuo2022deep}. Thus, classical RL is extended with model learning, where both $\theta$ and $\pi$ are optimized so that $\pi$ performs well in the real environment while exploiting the model for foresight.
This paradigm, where the agent learns the environment and plans based on the learned model, essentially falls under the umbrella of model-based RL (MBRL) research \cite{sutton1991dyna}. However, classical MBRL and modern world-model methods share the same ``learn-a-model-then-plan" spirit, yet differ significantly in their state representation, imagination capabilities, and approaches to uncertainty handling. Section~\ref{subsec:mbrl} provides a detailed side-by-side analysis.

\subsection{Technologies in Realizing World Models}

\subsubsection{Models for Latent Encoding and Decoding}

The most distinct feature of a world model, compared to direct inference methods, is its ability to encode agent observations into a latent space, enabling efficient extraction and modeling of environmental dynamics. Common encoding architectures include Autoencoders (AE), VAE, and Vector Quantized-VAEs (VQ-VAE) \cite{chen2023auto}. AEs learn compact latent states by minimizing reconstruction error between inputs and outputs. VAEs extend this by introducing probabilistic latent variables, optimizing the evidence lower bound (ELBO) to balance reconstruction and regularization \cite{doersch2016tutorial}. VQ-VAEs discretize the latent space with a codebook, improving representation learning and generative capability \cite{van2017neural}.

\subsubsection{Models for Dynamics Modeling}

Once encoded, latent states are used to predict temporal evolution, enabling both analysis of past data and imagination of future outcomes. This typically employs temporal networks and generative models:

\begin{itemize}
\item \textbf{RNNs:} RNNs capture temporal dependencies via evolving hidden states, making them suitable for sequential modeling. Mixture-density RNNs \cite{ha2018world} handle multimodal dynamics, while RSSMs, such as Dreamer \cite{hafner2019dream}, integrate deterministic and stochastic components. They use sequence variational inference to reconstruct observations and rewards while minimizing KL divergence between posterior and prior latent states.

\item \textbf{Autoregressive Models:} These predict the next step conditioned on past observations. Transformer-based models excel in discretized environments, trained by maximizing next-step likelihood with teacher forcing \cite{khan2022transformers}. They have been widely applied to language, where LLMs simulate textual ``worlds”, and to video prediction using image patches or visual tokens \cite{han2022survey}.

\item \textbf{Pre-trained Transformers:} Large multimodal transformers implicitly capture environmental structures. LLMs such as GPT \cite{yenduri2024gpt} generate coherent text scenarios, while vision-language and large vision models (VLMs, LVMs) \cite{zhang2024vision, liu2024sora} learn object interactions. DeepMind’s Gato \cite{reed2022generalist} unifies text, vision, and robotics into a single sequence model, serving as a generalist world model.

\item \textbf{Generative Adversarial Network (GANs):} GANs generate realistic outputs through an adversarial process between a generator and discriminator \cite{goodfellow2020generative}. Applied to world models, they capture future observation distributions \cite{awiszus2021world}, but training instability and mode collapse remain challenges \cite{creswell2018generative}.

\item \textbf{Diffusion Models:} Diffusion models generate data by iteratively denoising noise samples \cite{yang2023diffusion}. They offer flexible, temporally consistent sequence generation for world simulation \cite{ding2024diffusion}. OpenAI’s Sora \cite{brooks2024video} exemplifies this by producing coherent minute-long videos simulating complex physical dynamics.
\end{itemize}

To facilitate intuitive understanding, we provide a detailed comparison in Table~\ref{tab:dynamic_model_compare}, highlighting the key distinctions between different dynamics models.
Building on this foundation, the following sections delve into representative applications of world models across various domains, illustrating their versatility and effectiveness in complex tasks.

\begin{table*}[htp] \scriptsize
  \centering
  \caption{Comparison of different models for dynamics modeling in world models}
  \label{tab:dynamic_model_compare}
    \begin{tabular}{m{0.12\textwidth}<{\centering}||>{\raggedright\arraybackslash}m{0.12\textwidth}|m{0.10\textwidth}<{\centering}|m{0.20\textwidth}<{\centering}|m{0.15\textwidth}<{\centering}|m{0.15\textwidth}<{\centering}}
      \hline
      \textbf{Model family}  &  \textbf{Network Type} & \textbf{Data Representation} & \textbf{Strengths} & \textbf{Limitations} & \textbf{Algorithms} \\
       \hline
      \multirow{1}{0.09\textwidth}[0pt]{\centering RNNs} &
      MDN-RNN, RSSM & Continuous latent vectors of observations and rewards 
& \begin{itemize}[leftmargin=*]
      \item[\textcolor{green}{\ding{51}}] Lightweight, runs on edge devices
          \item[\textcolor{green}{\ding{51}}] Data-efficient
\item[\textcolor{green}{\ding{51}}] Differentiable multi-step roll-outs
      \vspace{-1.0em}
      \end{itemize}& \begin{itemize}[leftmargin=*]
      \item[\textcolor{red}{\ding{55}}] Vanishing gradients on long horizons
          \item[\textcolor{red}{\ding{55}}] Compounding prediction error during roll-outs
      \vspace{-1.0em}
      \end{itemize} 
      & \begin{itemize}[leftmargin=*]
      \item[\textcolor{blue}{\ding{108}}] World Models \cite{ha2018world}
          \item[\textcolor{blue}{\ding{108}}] PlaNet \cite{hafner2019learning}
        \item[\textcolor{blue}{\ding{108}}] Dreamer family \cite{hafner2019dream, hafner2020mastering, hafner2023mastering}
      \vspace{-1.0em}
      \end{itemize}
      \\
      \hline
        \multirow{1}{0.09\textwidth}[0pt]{\centering Autoregressive Transformers} &
      Transformer-XL, VQ-VAE, Transformer
      & Discrete token sequences & \begin{itemize}[leftmargin=*]
      \item[\textcolor{green}{\ding{51}}] Captures long temporal context
          \item[\textcolor{green}{\ding{51}}] Parallelizable training
          \item[\textcolor{green}{\ding{51}}] high sample fidelity
      \vspace{-1.0em}
      \end{itemize}& \begin{itemize}[leftmargin=*]
      \item[\textcolor{red}{\ding{55}}] Tokenisation may discard fine details
          \item[\textcolor{red}{\ding{55}}] Heavy compute demand
      \vspace{-1.0em}
      \end{itemize} & \begin{itemize}[leftmargin=*]
      \item[\textcolor{blue}{\ding{108}}] TWM \cite{robine2023transformer}
          \item[\textcolor{blue}{\ding{108}}] VideoGPT \cite{yan2021videogpt}
      \vspace{-1.0em}
      \end{itemize}
      \\
      \hline
      \multirow{1}{0.09\textwidth}[0pt]{\centering Pre-trained Transformer} &
      Large multimodal Seq-to-Seq Transformer, LLMs & Unified token streams & \begin{itemize}[leftmargin=*]
      \item[\textcolor{green}{\ding{51}}] Broad zero/ few-shot generalizationacross domains
          \item[\textcolor{green}{\ding{51}}] Convenient via public APIs
      \vspace{-1.0em}
      \end{itemize}& \begin{itemize}[leftmargin=*]
      \item[\textcolor{red}{\ding{55}}] Large model size
          \item[\textcolor{red}{\ding{55}}] Substantial data requirements
      \vspace{-1.0em}
      \end{itemize} & \begin{itemize}[leftmargin=*]
      \item[\textcolor{blue}{\ding{108}}] DeepMind Gato \cite{reed2022generalist}
          \item[\textcolor{blue}{\ding{108}}] Worldgpt \cite{ge2024worldgpt}
      \vspace{-1.0em}
      \end{itemize}\\
      \hline
      \multirow{1}{0.09\textwidth}[0pt]{\centering GANs} &
      Generator, Discriminator & Raw pixels or voxels of video frames & \begin{itemize}[leftmargin=*]
      \item[\textcolor{green}{\ding{51}}] High-resolution one-shot predictions
          \item[\textcolor{green}{\ding{51}}] Good spatial realism
      \vspace{-1.0em}
      \end{itemize}& \begin{itemize}[leftmargin=*]
      \item[\textcolor{red}{\ding{55}}] Mode collapse
          \item[\textcolor{red}{\ding{55}}] Unstable training
          \item[\textcolor{red}{\ding{55}}] Poor temporal coherence
      \vspace{-1.0em}
      \end{itemize} & \begin{itemize}[leftmargin=*]
      \item[\textcolor{blue}{\ding{108}}] World-GAN \cite{awiszus2021world}
          \item[\textcolor{blue}{\ding{108}}] DVD-GAN \cite{clark2019adversarial}
      \vspace{-1.0em}
      \end{itemize}\\\hline
      \multirow{1}{0.09\textwidth}[0pt]{\centering Diffusion Models} &
      Denoising Diffusion, Diffusion Transformer & Video frames or full action–state trajectories & \begin{itemize}[leftmargin=*]
      \item[\textcolor{green}{\ding{51}}] Stable likelihood-based training
          \item[\textcolor{green}{\ding{51}}] High visual fidelity
          \item[\textcolor{green}{\ding{51}}] Long-range consistency
      \vspace{-1.0em}
      \end{itemize}& \begin{itemize}[leftmargin=*]
      \item[\textcolor{red}{\ding{55}}] Slow sampling
          \item[\textcolor{red}{\ding{55}}] Large compute footprint
      \vspace{-1.0em}
      \end{itemize} & \begin{itemize}[leftmargin=*]
      \item[\textcolor{blue}{\ding{108}}] OpenAI Sora \cite{brooks2024video}
          \item[\textcolor{blue}{\ding{108}}] PolyGRAD \cite{rigter2023world}
      \vspace{-1.0em}
      \end{itemize}\\
      \hline
    \end{tabular}
\end{table*}

\section{Imagination-Driven Planning and Policy Optimization}


World models are designed to capture the latent dynamics that govern the evolution of an environment.
Once these dynamics are learned, they enable three primary downstream capabilities:
\textit{(i)} policy learning, where imagined rollouts can replace costly real-world interactions;
\textit{(ii)} high-fidelity video prediction and generation, which facilitates long-horizon forecasting of visual scenes; and
\textit{(iii)} structural reasoning for causal analysis.
Among these, we primarily focus on \emph{imagination-driven policy optimization}, as it most clearly demonstrates the potential of world models to support EGI optimizations.

\subsection{The Emergence of World Models}
\label{subsec:mbrl}

The concept of world models in AI originates from early MBRL. In the 1990s, Sutton’s Dyna architecture \cite{sutton1991dyna} demonstrated that learning an internal model of environment dynamics enables agents to simulate “imagined” experiences, improving planning and sample efficiency. Other works, such as prioritized sweeping \cite{moore1993prioritized}, optimized model updates, and predictions. These classical methods showed that even simple transition models can outperform purely model-free approaches. In wireless network design, coarse models such as path-loss formulas or stochastic geometry abstractions
have been used offline to compute control laws that later operate in real time \cite{zappone2019wireless}.

Building on these ideas, recent studies employ MBRL for wireless optimization. In \cite{imanberdiyev2016autonomous}, a quadcopter controller combined the TEXPLORE MBRL algorithm \cite{hester2013texplore} with a random-forest transition model learned from a few flights in Gazebo. Monte Carlo tree search rollouts refined the policy online, and the agent achieved energy-aware goal-seeking behavior with real-time execution, outperforming Q-learning. Beyond UAV trajectory planning, \cite{jafarzadeh2022model} proposed a multi-agent MBRL framework that integrates fuzzy-logic link assessment, where the fuzzy module provides transition probabilities to anticipate mobility-induced topology changes. This design improved packet delivery, reduced delays, and lowered signaling overhead by 5\% compared with model-free baselines. For network slicing, \cite{alcaraz2022model} introduced kernel-based RL (KBRL), where a lightweight classifier predicts SLA violations and guides resource allocation. KBRL reduced SLA-violation episodes by up to three times and halved computational time compared with deep model-free methods.
Despite these achievements, early MBRL in wireless networks relied on tabular or linear function approximations, which restricted their application to relatively small-scale environments \cite{zappone2019wireless}.

\begin{figure}[t]
    \centering
    \includegraphics[width= 0.95\linewidth]{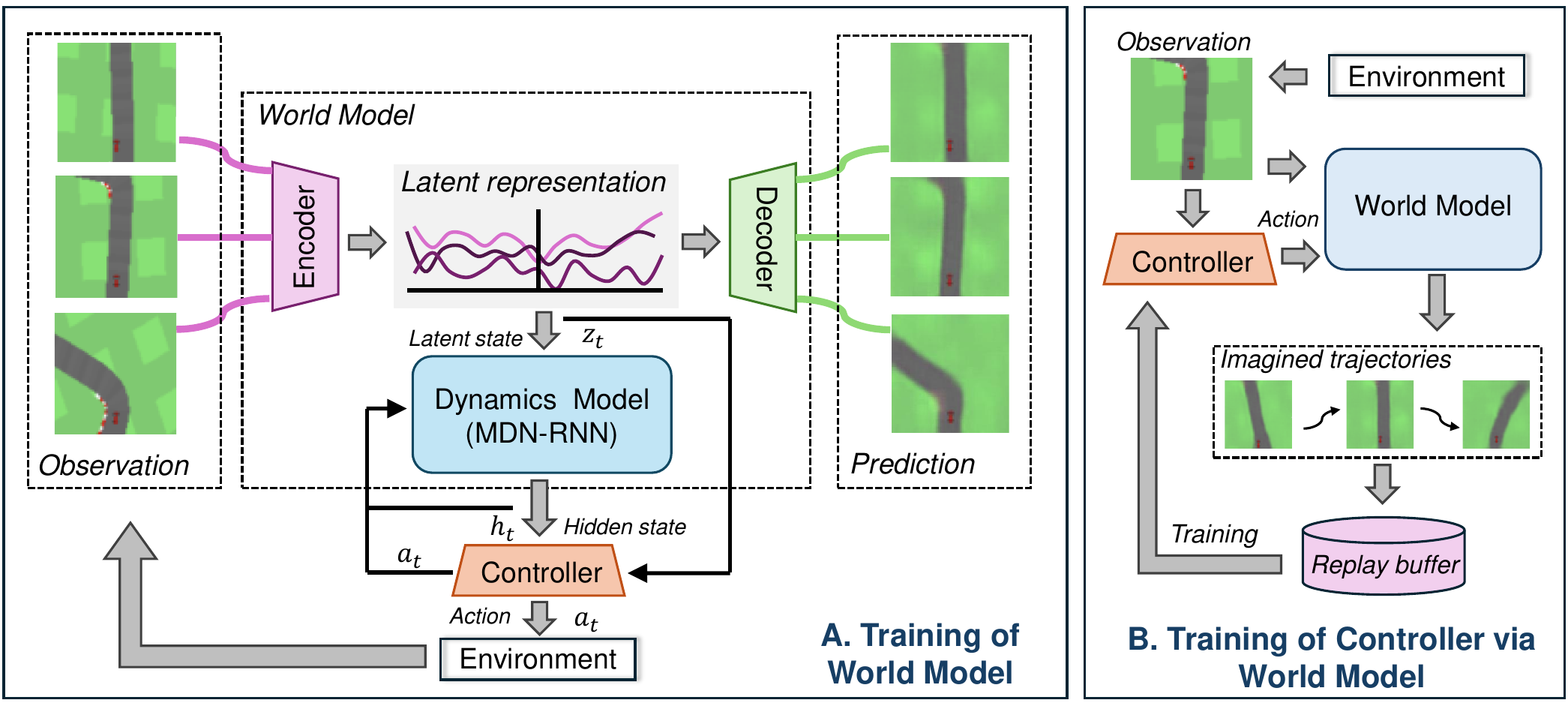}
    \caption{Flow diagram of world model in CarRacing \cite{ha2018world}. 
    \textit{Part A:} training process of world model. The observation is first processed by the encoder to produce the latent state $z_t$. Then, the dynamics model will transfer the latent state to the hidden state $h_t$ with action $a_t$. Finally, the controller will output the action $a_t$ based on the current hidden state $h_t$ and latent state $z_t$.
    \textit{Part B:} training process of the controller via the world model. 
    After observing the state of the environment, the world model will generate possible future trajectories and add them to the replay buffer. The controller is trained using the imagined trajectories without interacting with the real environment.
    }
    \label{fig:ex_world model}
\end{figure}

The resurgence of deep learning around 2015 sparked new progress in world models.
Researchers began training high-capacity neural networks to predict environment dynamics from high-dimensional inputs, enabling planning in complex domains. 
These world models learn an abstract, compact representation of the environment’s dynamics, enabling agents to plan and make decisions without relying on pixel-level predictions \cite{schrittwieser2020mastering}. Instead of reconstructing raw observations, these models focus on predicting task-relevant quantities, including future rewards, values, or abstract states \cite{ha2018world}.
By avoiding a full pixel reconstruction, encoder models drastically reduce the information they must capture, freeing capacity to represent only what is relevant for planning \cite{hafner2019learning}. In other words, the learned latent state does not need to match the true environment state or reconstruct observations, as long as it encodes the features necessary to predict future outcomes for decision-making \cite{ha2018world}. This approach has yielded powerful agents that ``imagine” future trajectories in a learned latent space, leading to improved sample-efficiency and performance in complex tasks.

To be more specific, the seminal work \cite{ha2018world} offers a concise yet powerful blueprint for what a world model can achieve, as shown in Fig. \ref{fig:ex_world model}. Using the \texttt{CarRacing-v0} benchmark, the authors first compress raw $64\times 64$ video frames into a compact latent vector with a VAE, and then train a recurrent mixture-density network (MDN-RNN) as the dynamics model to forecast how this latent representation will evolve under a sequence of steering, throttle, and brake commands. Once learned, the joint model operates as an internal simulator, which can predict hundreds of future steps without invoking the computationally heavy physics engine, allowing an external controller to evaluate candidate action sequences entirely offline.

Although CarRacing concerns a single vehicle on a procedurally generated track \cite{brockman2016openai}, the underlying principles in CarRacing and other video games translate directly to edge environments populated by mobile robots, UAVs, and heterogeneous sensors in EGI.
\begin{itemize}
    \item By predicting latent dynamics several seconds ahead, the agent gains \emph{temporal foresight}, which is the same sort of predictive power that an edge node would anticipate channel fades or sudden traffic surges~\cite{becvar2024machine, li2024deep}.
    \item The system supports \emph{long-horizon action-sequence search} by evaluating thousands of imagined trajectories, enabling the selection of paths that are simultaneously energy-efficient and safety-aware. This is an ability valuable to UAV networks and autonomous ground vehicular networks \cite{song2024energy, wu2021joint, liu2022joint}.
    \item The probabilistic nature of the RNN model makes it straightforward to sample diverse futures and quantify risk. This \emph{uncertainty-aware imagination} aligns with the need to assess rare but catastrophic events, such as wireless outages during disaster response \cite{wu2022uncertainty, xu2022uncertainty}. 
    \item The model can generate virtually unlimited sensor traces, providing \emph{synthetic data augmentation} in scenarios where collecting privacy-sensitive or safety-critical real data is impractical \cite{luo2024rm, wang2025generative, wang2024generative}.
\end{itemize}

\begin{figure*}[t]
    \centering
    \includegraphics[width= 0.95\linewidth]{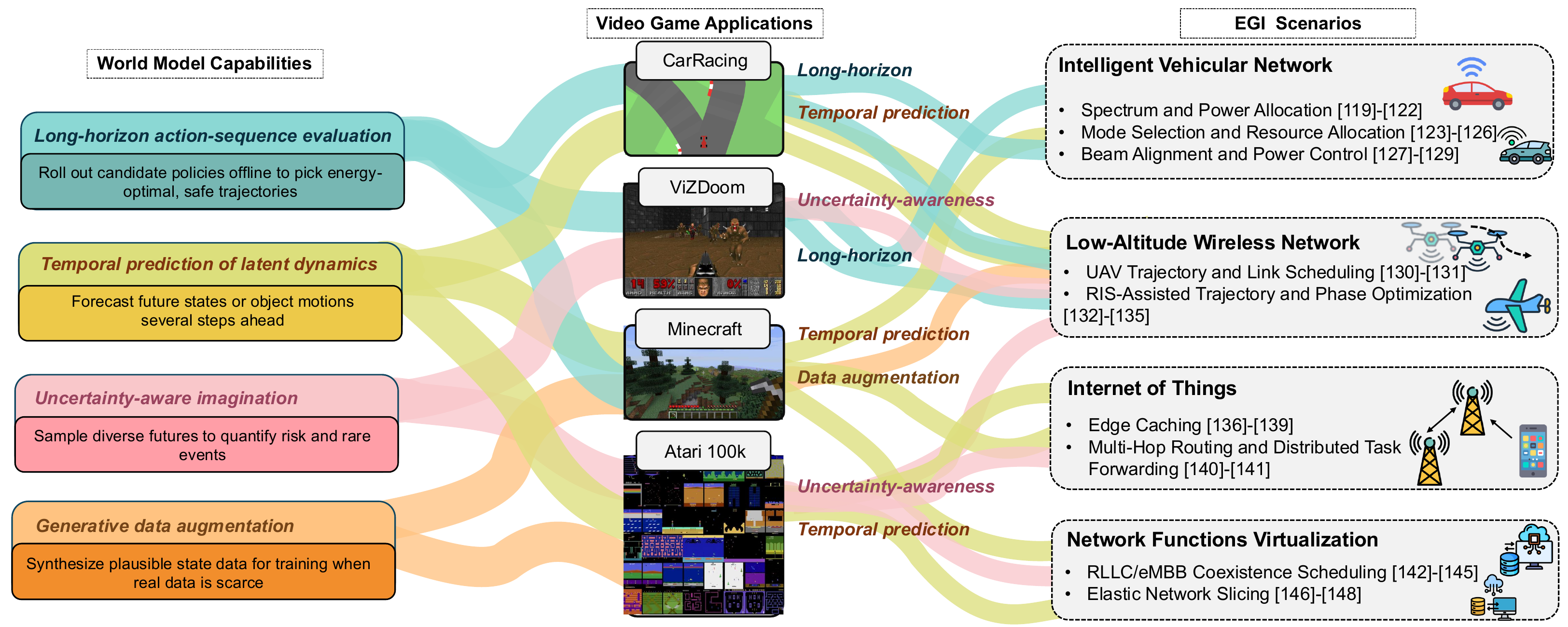}
    \caption{Mapping world model capabilities to video game scenarios and EGI applications.
Four key capabilities of world models are illustrated through video game benchmarks and mapped to EGI scenarios. These capabilities enable proactive planning and adaptive optimization in edge networks.
    }
    \label{fig:video_game}
\end{figure*}

In summary, a world model serves as an internal world simulation engine that resides on the edge. In this regard, CarRacing and other video games are more than a toy experiment. It is a microcosm of how predictive latent models can furnish foresight, risk awareness, and data efficiency, the very hallmarks of EGI. For a comprehensive illustration, we show the connection between video games and EGI applications in Fig. \ref{fig:video_game}.

\subsection{World Model Paradigm}

In 2018, Ha and Schmidhuber introduced the term “world models” for an RL agent that learns a generative model of its environment to complete video games \cite{ha2018world}. A key feature of this paradigm is the use of a VAE-based encoder to model the environment’s latent state as a continuous vector capturing spatial and temporal features. A separate controller is then trained to maximize rewards using only the latent state \cite{ha2018world}.

The idea of using AE or VAE to compress and extract essential features has also been widely applied in wireless communication and EGI networks \cite{sun2025generative}. Their inherent compress-and-reconstruct capability has proven effective in channel estimation tasks \cite{kim2023deep}. For example, the framework in \cite{kim2023deep} achieves the lowest mean squared error while requiring 20\% fewer pilot symbols than conventional methods. In \cite{mohamed2019model}, an end-to-end AE extracts input features, achieving bit error rate performance comparable to traditional modulation while reducing system complexity. Another study \cite{sun2025generative} integrates a VAE into DRL to extract latent representations from high-dimensional sensor data, improving sample efficiency and policy robustness. In their UAV-assisted communication case study, VAE-based dimensionality reduction lowers state space complexity while preserving critical information.
Although existing works in EGI networks mainly use VAEs for state compression or feature extraction, they remain limited to preprocessing roles rather than being integrated into the decision-making pipeline \cite{sun2025generative}. In contrast, the controller in world models is trained entirely within the learned model without additional real environment interaction, allowing the agent to “practice by prediction” \cite{ha2018world}.

The world models in \cite{ha2018world} used an open-loop controller optimized offline rather than a closed-loop planner at decision time. Building on this, PlaNet introduced online planning with latent dynamics~\cite{hafner2019learning}. PlaNet learns an RSSM with deterministic and stochastic latent components to capture multiple possible futures. Using model-predictive control (MPC)\cite{garcia1989model} in latent space, it simulates many candidate action sequences and selects the one with the highest predicted cumulative reward at each step. This approach allows strong performance with about 200 times less environment interaction than model-free methods\cite{hafner2019learning}. Because both planning and reward prediction operate entirely in latent space, PlaNet avoids generating images during search, enabling efficient evaluation of thousands of trajectories in parallel~\cite{hafner2019learning}. Similar latent-space MPC methods have been applied to EGI optimization tasks, including control, power transfer, and wireless charging \cite{minavrik2024model, zhou2020model, ma2024low}, but they often rely on task-specific dynamics or handcrafted costs rather than fully self-supervised models.

Another research line combines latent models with tree search, exemplified by DeepMind’s MuZero~\cite{schrittwieser2020mastering}. MuZero’s model has a representation function encoding history into a latent state, a dynamics function predicting the next state, and heads estimating reward, value, and policy. Based on the current latent state, Monte Carlo Tree Search (MCTS)\cite{browne2012survey} explores possible action sequences, using predicted rewards and values to estimate returns. MCTS supports effective decision-making in uncertain scenarios with low computational cost, making it suitable for UAV trajectory optimization \cite{qian2022path}. By focusing on policy and value prediction rather than pixel reconstruction, MuZero allows the latent state to encode any internal logic required for planning. On games such as Go, chess, and shogi, it achieved superhuman performance comparable to AlphaZero\cite{silver2017mastering}, despite lacking explicit game rules~\cite{schrittwieser2020mastering}. This design avoids irrelevant details, reduces complexity, and integrates the strengths of model-based and model-free learning for efficient lookahead planning.

\subsection{Learning Policies inside Latent Dynamics}

\begin{figure}[t]
    \centering
    \includegraphics[width= 0.80\linewidth]{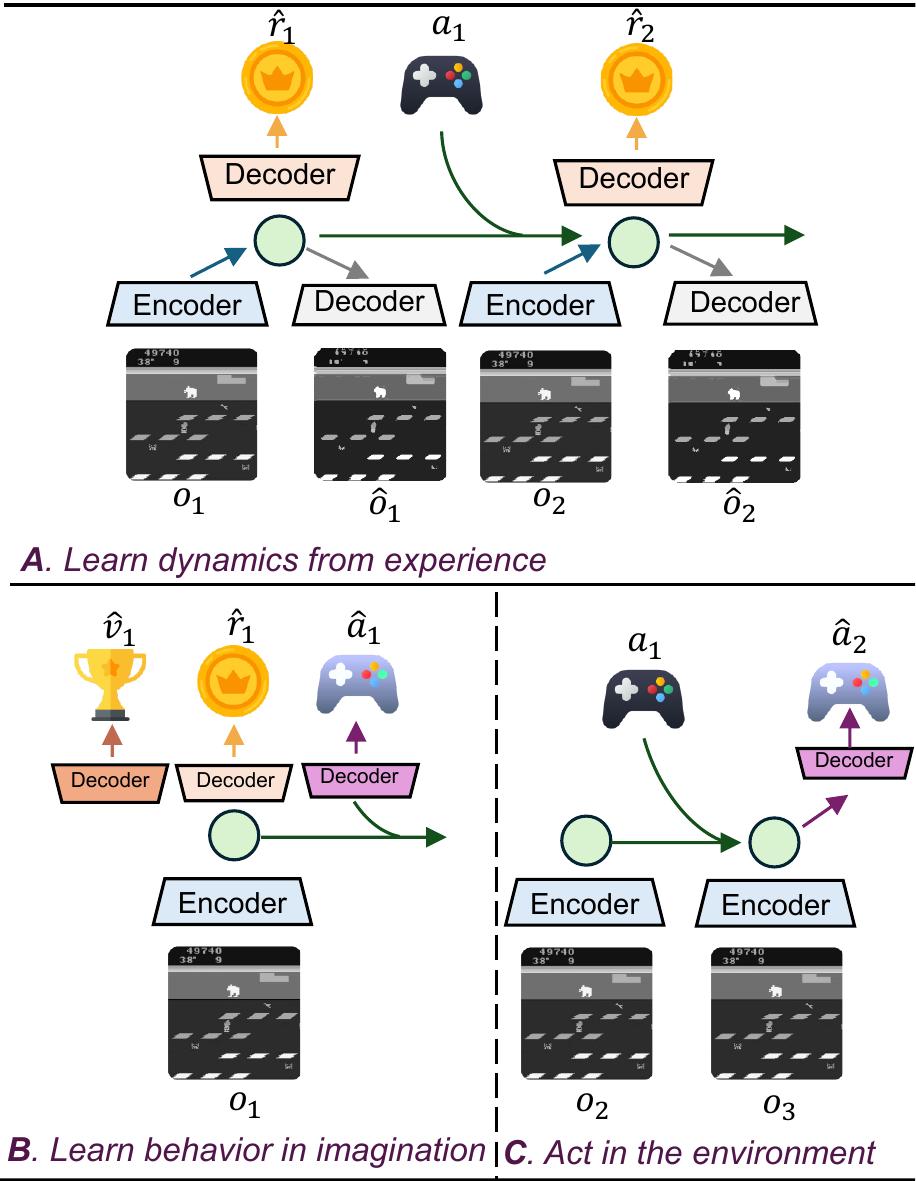}
    \caption{Components of Dreamer. \textit{Part A:} Agent learns to encode observations $o_i$ and actions into compact latent states via reconstruction $\hat{o}_i$ and predicts
environment rewards $\hat{r}_i$. \textit{Part B:} Dreamer predicts state values $\hat{v}_i$ and actions $\hat{a}_i$ in the latent space. \textit{Part C:} The agent encodes the history of the episode to compute the current model state and
predict the next action.}
    \label{fig:dreamer}
\end{figure}

Rather than planning at each step, the Dreamer family of methods~\cite{hafner2019dream, hafner2020mastering, hafner2023mastering} trains a policy network entirely through “imagination.” Dreamer applies an actor-critic approach that optimizes long-horizon behaviors within the latent space of a learned dynamics model, as illustrated in Fig. \ref{fig:dreamer}. Its world model is an RSSM, which combines a deterministic RNN core with stochastic latent variables.
While classical RNNs have been applied to UAV angle and CSI prediction \cite{yuan2020learning}, Dreamer introduces two key innovations. First, the RSSM is variationally trained to reconstruct both observations and rewards, producing a belief over future trajectories rather than a single forecast. Second, once trained, Dreamer rolls out imagined trajectories in latent space and learns an actor and critic by backpropagating through the model without further environment interaction \cite{hafner2019dream}.
The agent predicts future latent trajectories, evaluates $n$-step returns using the learned reward model, and updates the value function toward these returns while improving the policy to maximize value estimates. Dreamer’s value network bootstraps rewards beyond the horizon of imagination, reducing the short-sightedness of finite-horizon planning. After $5\times 10^6$ environment steps, Dreamer achieves an average score of 823, outperforming 332 of PlaNet and surpassing the best model-free baseline D4PG, which achieves 786 after $10^8$ steps \cite{hafner2019dream}.

Although Dreamer achieved state-of-the-art performances on various control tasks, 
Dreamer and related algorithms typically maintain continuous latent states and optimize the policy with gradient-based methods. 
An evolution of this line is DreamerV2, which introduced a discrete latent space for the world model \cite{hafner2020mastering}.
DreamerV2 represents each latent state with a set of categorical variables instead of a Gaussian in Dreamer, which helps capture multimodal uncertainty in dynamics.
DreamerV2 jointly trains the world model and policy from pixel inputs without reconstructing images accurately, relying instead on latent dynamics consistency and reward prediction. This encourages the model to focus its capacity on features that are relevant to future rewards. Notably, DreamerV2 became the first model-based agent to achieve human-level performance on the Atari benchmark with 55 games by learning behaviors purely inside its learned world model \cite{hafner2020mastering}.


Several follow-up studies have further refined world model training to improve policy learning.
In \cite{deng2022dreamerpro}, DreamerPro replaced pixel-level reconstruction with a contrastive clustering objective. The world model is trained to map each observation to one of a small set of prototype latent codes, or ``proto-states”, and to predict rewards directly in that discrete latent space. 
A parallel line of research in remote-sensing of EGI has confirmed the broader appeal of this prototype-driven abstraction \cite{ma2023multipretext, wang2023multiscale}.
MPCL-Net leveraged self-supervised scale and rotation pretext tasks to derive multiview features and then anchors them to class-level prototypes,
enabling few-shot scene classification with very limited labels \cite{ma2023multipretext}.
MPCNet goes further for high-resolution aerial imagery segmentation. It extracted scene-specific multiscale prototypes 
and applies a multiscale prototype-contrastive loss to suppress intraclass heterogeneity across scenes \cite{wang2023multiscale}. 
Both works echo DreamerPro’s insight that a compact, semantically meaningful prototype space can simultaneously dispense with heavy pixel reconstruction and improve sample efficiency of imagination or dense prediction.
In the DeepMind Control (DMC) suite \cite{tassa2018deepmind}, DreamerPro achieves performance comparable to Dreamer under standard settings. However, when the visual background is replaced with real-world scenes, significantly increasing the reconstruction difficulty, DreamerPro outperforms Dreamer by more than fivefold \cite{deng2022dreamerpro}.
This reconstruction-free approach avoids the issue of the model wasting capacity on modeling irrelevant visual details, similar in spirit to MuZero \cite{schrittwieser2020mastering}. 

Similarly, Dreaming \cite{okada2021dreaming} and its successor DreamingV2~\cite{okada2022dreamingv2} likewise aim to train world models ``without reconstruction”.
Dreaming combines a contrastive loss on latent predictions with the Dreamer framework, so that the world model is trained to predict future latent states that are distinguishable from random distractors rather than to reconstruct images. DreamingV2 builds on this by using discrete latents together with contrastive learning. On the 3D robotic arm tasks, DreamingV2 outperforms DreamerV2 by over 30\%, showing improved performance compared to models trained with autoencoder losses \cite{okada2022dreamingv2}.
The motivation for these methods is that autoencoding pixels can lead to ``object vanishing” and sensitivity to visual distractions, such as the model might ignore small task-relevant features or overfit to background noise \cite{okada2021dreaming}. By forgoing pixel reconstruction, the latent model is encouraged to encode only the information needed to predict rewards and high-level state transitions, which can improve robustness and generalization.

The latest model in the Dreamer family is DreamerV3~\cite{hafner2023mastering}. 
DreamerV3 builds upon its predecessors by significantly enhancing generalizability and robustness across diverse tasks. It integrates robust techniques including normalization, balancing, and transformations, enabling stable learning with fixed hyperparameters across more than 150 tasks, including Atari games, ProcGen, DMLab, continuous control, and even the complex Minecraft environment \cite{hafner2023mastering}. Notably, DreamerV3 is the first reinforcement learning algorithm capable of autonomously achieving the challenging task of collecting diamonds in Minecraft from scratch without any human data or task-specific heuristics \cite{hafner2023mastering}. 


In summary, the Dreamer family of methods all use actor-critic learning on imagined latent trajectories, which provides several options for EGI optimization:
\begin{itemize}
    \item \textbf{Sample-Efficient Training:}  Latent models optimize wireless behaviors using far fewer environment interactions, crucial for data-limited wireless scenarios.
    \item \textbf{High-Dimensional Channel Modeling:} It scales effectively to complex wireless environments by leveraging expressive, discrete latent representations.
    \item \textbf{Cross-Environment Generalization:} Minimal domain-specific tuning enables deployment across diverse wireless conditions and network topologies.
\end{itemize}

\subsection{Constrained World Models}

As research progressed, new variants of the world model have addressed longer-term dependencies, safety considerations, and resource constraints for EGI \cite{chen2022transdreamer, huang2023safedreamer,chun2025sparse,robine2025simple}. 


Inspired by the longer temporal dependencies of transformers \cite{vaswani2017attention}, the authors proposed a Transformer State-Space Model (TSSM), a stochastic latent dynamics model implemented with transformer blocks, which replaces the RNN-based dynamics model with a transformer model in a latent state-space~\cite{chen2022transdreamer}.
The TSSM is trained similarly to Dreamer’s RSSM, but it can be parallelized and can maintain longer context \cite{chen2022transdreamer}. 
The same mechanism underpins a recent transformer-based RL framework for UAV area coverage and AoI-aware data collection \cite{chen2024transformer, zhu2022uav}. 
By re-weighting a variable-length set of neighbor observations, transformer structure coordinates flight paths that lift the average‐coverage score by 40\% and improve the fairness index by 45\% over MLP baselines~\cite{chen2024transformer}.
In the challenging 4-Ball configuration, TransDreamer significantly outperforms Dreamer, achieving a success rate of 23\%, whereas Dreamer reaches only 7\%~\cite{chen2022transdreamer}. 
This demonstrates that a transformer-based latent model can imagine further into the future or handle long sequences without forgetting.

Another important direction is ensuring safety in EGI planning. Standard world model agents optimize for reward, which can lead to unsafe behavior if there are hidden costs not captured by the reward, such as unknown attacks. SafeDreamer augments the Dreamer framework with constrained planning to respect safety constraints during decision-making \cite{huang2023safedreamer}. 
SafeDreamer integrates a safety signal into the world model and uses a Lagrangian optimization approach to balance reward and safety during planning.
They utilized the constrained cross-entropy method \cite{wen2018constrained} during imagination to filter out action sequences that would likely result in constraint violations. 
By planning within the learned model for no-violation trajectories, SafeDreamer achieved nearly zero safety violations on complex vision-based tasks in the Safety Gymnasium benchmark, while still accomplishing the task goals \cite{huang2023safedreamer}. 
This integration of dreaming and planning enables the agent to anticipate and avoid unsafe outcomes before they materialize in the real world, such as steering a UAV away from high-interference channels or regions with signal blockages during a communication task.

Recent research increasingly considers practical constraints such as computational efficiency and model complexity, which are critical for the deployment of agentic AI systems at the edge \cite{you2016energy}.
In \cite{chun2025sparse}, the authors propose Sparse Imagination, a visual world model designed explicitly for resource-constrained edge environments. Their method enhances computational efficiency by selectively processing visual tokens via randomized grouped attention within transformer-based dynamics models. This approach dramatically accelerates planning by significantly reducing the computational overhead required for imagining future trajectories, making it particularly suitable for real-time decision-making in resource-limited edge scenarios.
Sparse Imagination demonstrates high control fidelity across diverse visual control tasks. In the PushT environment, it reduces the episode duration from 173 seconds of baseline to 82 seconds, while maintaining a comparable success rate 78.3\%, where baseline is 75.0\%~\cite{chun2025sparse}. 

Another significant advancement toward simplifying world models for practical edge deployment is introduced by Robine et al. through their Simple, Good, Fast (SGF) model \cite{robine2025simple}. In contrast to many contemporary approaches, SGF eliminates the reliance on RNNs, transformers, and discretized representations. Instead, it adopts self-supervised representation learning in conjunction with straightforward frame and action stacking techniques. 
By explicitly prioritizing computational simplicity and robustness, SGF achieves notable performance improvements while significantly reducing both training and inference complexity. On the Atari 100k benchmark, SGF cuts total training time from 12 hours to just 3 hours for DreamerV3~\cite{robine2025simple}.
This class of lightweight yet effective world models aligns well with the demands of agentic AI systems deployed at the edge, where computational constraints are stringent, but fast and reliable decision-making is essential~\cite{robine2025simple}.

\begin{table*}[htp] \scriptsize
  \centering
  \caption{Comparison of World Model-based planning methods}
  \label{tab:planning}
    \begin{tabular}{m{0.09\textwidth}<{\centering}||>{\raggedright\arraybackslash}m{0.12\textwidth}<{\centering}|m{0.15\textwidth}<{\centering}|m{0.18\textwidth}<{\centering}|m{0.30\textwidth}}
      \hline
      \textbf{Method}  &  \textbf{Model Type} & \textbf{Latent Space} & \textbf{Planner} & \textbf{Key Contributions} \\
       \hline
      \multirow{1}{0.09\textwidth}[0pt]{\centering World Models \cite{ha2018world}} &
      VAE + RNN & Continuous (Gaussian VAE code + RNN state) & Policy evolved in the latent without online planning & 
      \begin{itemize}[leftmargin=*]
      \item[\textcolor{blue}{\ding{108}}] Introduced world models for RL
          \item[\textcolor{blue}{\ding{108}}] Learned compact visual and memory model, enabling agent training ``inside its dream”
      \vspace{-1.0em}
      \end{itemize}\\
      \hline
      \multirow{1}{0.09\textwidth}[0pt]{\centering PlaNet \cite{hafner2019learning}} &
      RSSM: stochastic + deterministic & Continuous (Gaussian posterior state) & MPC with cross-entropy method & 
      \begin{itemize}[leftmargin=*]
      \item[\textcolor{blue}{\ding{108}}] First purely latent-space MPC from pixels
          \item[\textcolor{blue}{\ding{108}}] RSSM captures multi-step uncertainty
      \vspace{-1.0em}
      \end{itemize}\\
      \hline
      \multirow{1}{0.09\textwidth}[0pt]{\centering MuZero \cite{schrittwieser2020mastering}} &
      Deep neural network with latent Markov decision process & Continuous hidden state (no explicit encoder-decoder) & MCTS tree search with learned policy & 
      \begin{itemize}[leftmargin=*]
      \item[\textcolor{blue}{\ding{108}}] Learned latent model that predicts rewards, values, and policies directly
          \item[\textcolor{blue}{\ding{108}}] Superhuman planning in Go/Chess/Atari with learned dynamics
      \vspace{-1.0em}
      \end{itemize}\\
      \hline
      \multirow{1}{0.09\textwidth}[0pt]{\centering Dreamer \cite{hafner2019dream}} &
      VAE encoder + RSSM & Continuous (stochastic latent with Gaussian) & Learned policy via actor-critic on imagined trajectories & 
      \begin{itemize}[leftmargin=*]
      \item[\textcolor{blue}{\ding{108}}] Introduced learning long-horizon behaviors by backpropagating through latent model
          \item[\textcolor{blue}{\ding{108}}] Data-efficiency in control tasks
      \vspace{-1.0em}
      \end{itemize}\\
      \hline
      \multirow{1}{0.09\textwidth}[0pt]{\centering DreamerV2 \cite{hafner2020mastering}} &
      RSSM with discrete latent codes & Discrete (multiple categorical latents) & Actor-critic in latent space & 
      \begin{itemize}[leftmargin=*]
      \item[\textcolor{blue}{\ding{108}}] Discrete latents to capture multimodal dynamics
          \item[\textcolor{blue}{\ding{108}}] Learned behaviors purely from latent predictions
      \vspace{-1.0em}
      \end{itemize}\\
      \hline
      \multirow{1}{0.09\textwidth}[0pt]{\centering DreamerPro \cite{deng2022dreamerpro}} &
      RSSM with prototypical representation & Discrete prototype vectors & Reconstruction-free actor-critic in latent space & 
      \begin{itemize}[leftmargin=*]
      \item[\textcolor{blue}{\ding{108}}] Removed reconstruction loss by learning prototype latent states
          \item[\textcolor{blue}{\ding{108}}] Improved robustness to irrelevant visual details
      \vspace{-1.0em}
      \end{itemize}\\
      \hline
      \multirow{1}{0.09\textwidth}[0pt]{\centering Dreaming \& DreamingV2 \cite{okada2021dreaming, okada2022dreamingv2}} &
      RSSM variants with contrastive learning & Continuous or Discrete & Actor-critic in latent space with contrastive prediction & 
      \begin{itemize}[leftmargin=*]
      \item[\textcolor{blue}{\ding{108}}] Pioneered contrastive reconstruction-free world model training
          \item[\textcolor{blue}{\ding{108}}] Avoided ``object vanishing” issue by focusing on latent prediction consistency
      \vspace{-1.0em}
      \end{itemize}\\
      \hline
      \multirow{1}{0.09\textwidth}[0pt]{\centering DreamerV3 \cite{hafner2023mastering}} &
      RSSM & Continuous (stochastic latent) & Actor-critic in latent space & 
      \begin{itemize}[leftmargin=*]
      \item[\textcolor{blue}{\ding{108}}] Demonstrated a single configuration mastering 150+ diverse tasks
          \item[\textcolor{blue}{\ding{108}}] Introduced robust normalization and objective balancing to generalize across domains
      \vspace{-1.0em}
      \end{itemize}\\
      \hline
      \multirow{1}{0.09\textwidth}[0pt]{\centering TransDreamer \cite{chen2022transdreamer}} &
      TSSM +  Transformer policy & Continuous (stochastic transformer embeddings) & Actor-critic in latent space with long-memory predictions & 
      \begin{itemize}[leftmargin=*]
      \item[\textcolor{blue}{\ding{108}}] First transformer-based world model for RL
          \item[\textcolor{blue}{\ding{108}}] Handled long-range dependencies better than RNN-based Dreamer
      \vspace{-1.0em}
      \end{itemize}\\
      \hline
      \multirow{1}{0.09\textwidth}[0pt]{\centering SafeDreamer \cite{huang2023safedreamer}} &
      RSSM + safety critic & Continuous (latent state with cost/reward predictions) & Actor-critic + online safe planning & 
      \begin{itemize}[leftmargin=*]
      \item[\textcolor{blue}{\ding{108}}] Integrated constrained RL with world models
          \item[\textcolor{blue}{\ding{108}}] Planned action sequences to minimize cost and maximize reward
      \vspace{-1.0em}
      \end{itemize}\\
      \hline
      \multirow{1}{0.09\textwidth}[0pt]{\centering Sparse Imagination \cite{chun2025sparse}} &
      Transformer world model that predicts future ViT patch tokens & Set of spatial patch tokens & MPC with cross-entropy method & 
      \begin{itemize}[leftmargin=*]
      \item[\textcolor{blue}{\ding{108}}] Introduces adaptive token dropout
          \item[\textcolor{blue}{\ding{108}}] Provided large runtime savings with near-full success on diverse manipulation
      \vspace{-1.0em}
      \end{itemize}\\
      \hline
      \multirow{1}{0.09\textwidth}[0pt]{\centering SGF \cite{robine2025simple}} &
      Feed-forward dynamics & Continuous (frame + action stacking) & Actor–critic on imagined latent predictions & 
      \begin{itemize}[leftmargin=*]
      \item[\textcolor{blue}{\ding{108}}] Showed that a minimal world model plus data augmentation \& stacking can reach competitive Atari-100k scores
          \item[\textcolor{blue}{\ding{108}}] Temporal consistency without added “baggage”
      \vspace{-1.0em}
      \end{itemize}\\
      \hline

    \end{tabular}
\end{table*}


\subsection{Summary and Comparison}

Table \ref{tab:planning} provides a comparative summary of representative world model-based planning methods, categorized by their model architecture, type of latent space, planning or control approach, and core contributions. Despite their differences, all these methods leverage latent predictive modeling to enable more efficient or effective decision-making than planning at the pixel level. By learning an internal model of the world’s dynamics, agents can reason about future consequences of actions. This paradigm has progressively evolved to integrate discrete generative abstractions, long-horizon memory through transformers, and considerations for safety and resource constraints, representing a substantial step toward realizing agentic AI within EGI.

\section{Using World Models to Enhance Optimization in Edge General Intelligence}

EGI optimization problems span diverse domains, each with unique objectives and constraints. 
Inspired by the excellent performance of world models, researchers have proposed a series of specialised variants for EGI deployment \cite{park2023model, wang2025world, zhao2025world}.

Targeting dense EGI deployments with partial observability, the authors in \cite{park2023model} embedded a neural partially observable Markov decision process (POMDP) world model into an access-point agent.
The learned world model jointly captures traffic arrivals and channel-occupancy dynamics, enabling Monte-Carlo planning to select contention actions without excessive real-world probing.
When compared to deep Q-learning baselines, the proposed framework can increase the radio success ratio by over 50\% while utilizing only 1/20 of the training data \cite{park2023model}.


Faced with bursty millimeter-wave (mmWave) blockages and ultra-short coherence times in vehicle-to-everything (V2X) links, the authors in \cite{wang2025world} coupled a world model with an actor-critic head so that long-horizon link-scheduling policies can be trained almost entirely in latent imagination.
By differentiating through imagined trajectories, the agent learns to attribute future packet-completeness-aware age-of-information (CAoI) to earlier decisions, boosting data efficiency and planning foresight. In a Sionna-based ray-tracing simulator, the framework trims CAoI by 26\% over model-based RL and 16\% over model-free RL, proving that learned environment simulators can keep vehicular networks fresh even during sensing outages~\cite{wang2025world}.

Another visionary paper \cite{zhao2025world} positioned the world model as the ``cognitive engine” inside autonomous edge agents. The authors devised Wireless Dreamer, a Dreamer-style latent model that plans UAV trajectories under weather uncertainty. A case study on weather-aware LAWNs shows that the model augments missing perception data and delivers higher-quality decisions than conventional optimization, where 
Wireless Dreamer accelerates convergence by 46.15\% compared to traditional DQN \cite{zhao2025world}.
This demonstrates the potential that how imagination can replace cloud-side optimization loops in resource-constrained edge hardware.

Taken together, these advances demonstrate that the world model paradigm can be systematically adapted to deliver safer, farther-looking, and resource-efficient decision-making, which are indispensable for next-generation EGI systems.
In the following, we review how world models can enhance optimization in four representative edge scenarios (Fig. \ref{fig:video_game}) by utilizing internal predictive models of the environment.






\subsection{Intelligent Vehicular Network}

\begin{figure}[t]
    \centering
    \includegraphics[width= 0.95\linewidth]{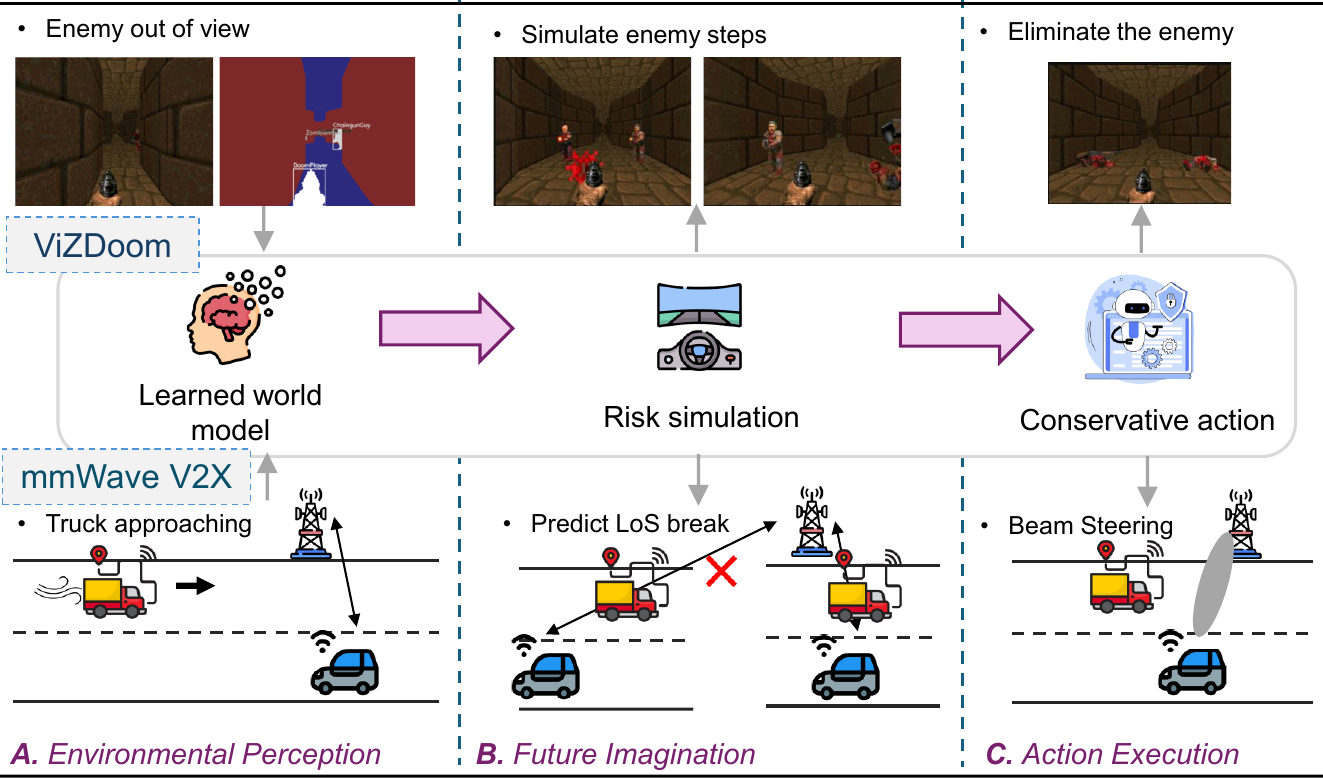}
    \caption{A conceptual illustration comparing ViZDoom gameplay with mmWave V2X communication. \textit{Part A, Environmental perception:} the agent detects threats such as an out-of-view enemy or an approaching truck. \textit{Part B, Future imagination:} the learned world model simulates possible risks, including enemy movement or LoS blockage. \textit{Part C, Action execution:} the agent takes conservative actions, such as shooting or beam steering, based on risk-aware predictions.}
    \label{fig:game2}
\end{figure}


\subsubsection{Spectrum and Power Allocation}

In vehicular communication (V2X) networks, the joint allocation of spectrum resources and transmission power represents a critical optimization challenge. The primary objective is to maximize system throughput while ensuring quality-of-service (QoS) requirements for safety-critical applications and managing co-channel interference in spectrum-sharing scenarios \cite{wang2022joint}. 
This NP-hard mixed-integer nonlinear programming (MINP) problem is further complicated by high mobility in vehicular environments, causing channel uncertainty and delayed CSI~\cite{wang2023joint}.
Recent solutions include RL-based distributed schemes \cite{wang2022joint}, robust optimization under imperfect CSI \cite{wang2023joint}, and multiaccess resource matching \cite{ibrahim2024optimizing}.
These traditional methods often rely on reactive approaches that respond to current or outdated network state information.

World models can solve V2X spectrum-power optimization due to their capability for long-horizon action-sequence evaluation and temporal prediction of latent dynamics. Instead of myopic allocations at each time instant, a world model-equipped agent could predict vehicle positions, channel quality, and interference levels over extended horizons, enabling proactive resource management. Just as agents in CarRacing learn track layouts and plan steering moves many turns in advance \cite{ha2018world}, V2X schedulers can anticipate future channel conditions and plan spectrum allocation sequences accordingly, rather than optimizing only immediate throughput. 
Similarly, the authors in \cite{sang2021deep} leverage predictive power allocation that achieves impressive prediction accuracy of 85.71\% and throughput performance remarkably close to optimal solutions under delayed CSI conditions.


\subsubsection{Mode Selection and Resource Allocation}

Another vehicular optimization problem in vehicular networks is joint mode selection and resource allocation in cellular V2X communications \cite{li2019joint}. Vehicles must choose between direct V2V sidelink or cellular uplink/downlink communication, while allocating radio resources such as frequency channels and time slots. The objective is to maximize network utility, such as throughput or connectivity, under strict latency and reliability constraints for safety messages~\cite{zhang2019deep, ye2019deep}.


Compared to continuous spectrum-power allocation, mode selection involves discrete decisions that reshape interference patterns \cite{ye2019deep}, making long-horizon action-sequence evaluation crucial. World models can be applied to this task, as they can anticipate future dynamics and evaluate the long-term impact of mode choices.
An edge controller could simulate scenarios such as ``If these 5 vehicles all choose direct V2V mode on channel X, what will the interference be 100 ms from now as they move closer?", thus avoiding unstable choices. Just as agents in CarRacing learn track layouts and plan steering moves many turns in advance \cite{ha2018world}, V2X schedulers can anticipate future channel congestion and plan spectrum allocation sequences accordingly. 
Similarly, the authors in \cite{han2021primary} employed a GAN to simulate the spectrum environment, enabling the system to effectively elude both primary user (PU) signals and jamming attacks while achieving faster convergence than traditional methods. Analogous to how agents in VizDoom anticipate incoming threats before they appear on screen \cite{ha2018world}, V2X agents can leverage imagination-based planning to foresee vehicle mobility patterns and interference scenarios, thereby selecting more robust communication modes in advance.

\subsubsection{Beam Alignment and Power Control}

In mmWave V2X communications, the optimization objective centers on joint beam alignment and power control to maximize spectral efficiency while maintaining low-latency connectivity under strict reliability constraints \cite{tan2024beam}. 
Recent advances \cite{rasheed2025deepbeam,ye2023beam} have demonstrated various approaches for mmWave beam management, and current methods primarily focus on immediate beam selection.

Similar to how agents in VizDoom (Fig. \ref{fig:game2}) learn to predict incoming projectiles several frames before they appear on screen by world models \cite{ha2018world}, mmWave agents can anticipate signal blockages and beam degradation by modeling the evolution of vehicle-blocker geometry over time \cite{rasheed2025deepbeam}. A learned world model could encode patterns such as ``adjacent truck will block line-of-sight in 0.5 seconds", predictions not immediately obvious from current received signal strength but inferable from motion trajectories.
The uncertainty-aware imagination capability proves crucial for handling the partial observability inherent in mmWave environments. Just as Minecraft agents must navigate uncertain terrain where obstacles may be hidden beyond their field of view \cite{hafner2023mastering}, mmWave agents operate with incomplete channel knowledge and must make beam decisions under uncertainty \cite{tan2024beam}. The world model can estimate confidence in its channel predictions during abrupt vehicle maneuvers, enabling the agent to take conservative actions, including beam widening or power boosting, when prediction confidence is low.

\begin{figure*}[t]
    \centering
    \includegraphics[width= 0.80\linewidth]{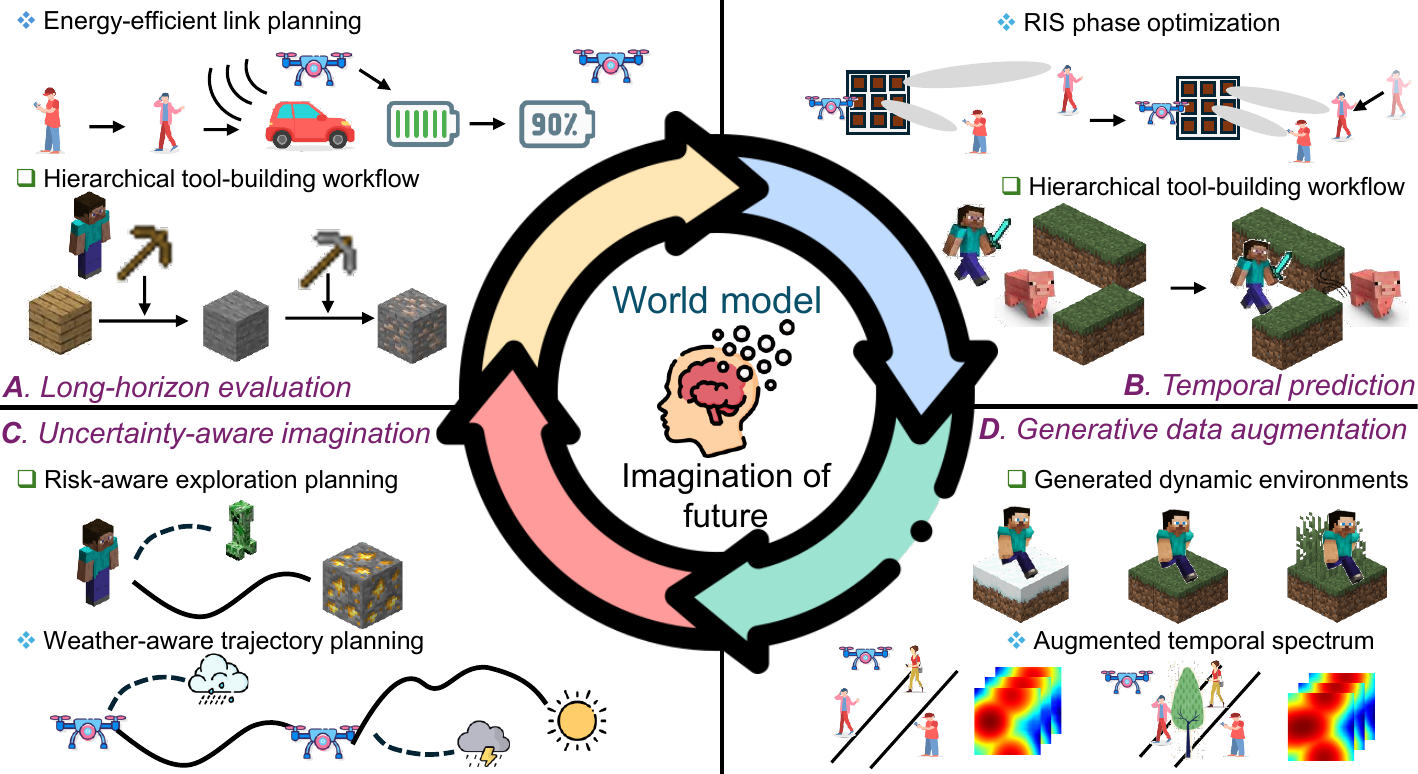}
    \caption{A conceptual illustration linking Minecraft gameplay with EGI tasks enabled by world models.
\textit{Part A, Long-horizon evaluation:} agents plan energy-efficient links or build tools by reasoning over extended action sequences.
\textit{Part B, Temporal prediction:} world models forecast dynamic changes, such as RIS-induced signal shifts or interactive environment evolution.
\textit{Part C, Uncertainty-aware imagination:} agents explore under risk, accounting for weather disruptions or hidden threats.
\textit{Part D, Generative data augmentation:} synthetic environments and spectral variations are produced to improve learning robustness.}
    \label{fig:game1}
\end{figure*}

\subsection{Low-Altitude Wireless Network}

\subsubsection{UAV Trajectory and Link Scheduling}

Low-altitude wireless networks (LAWNs) often deploy UAVs as mobile relays or base stations to provide flexible and high-quality communication coverage. 
The goal is typically to maximize the cumulative throughput or minimize mission completion time while satisfying UAV energy constraints and QoS requirements, such as per-user data rate guarantees or waypoint constraints \cite{shi2019multi}.

World models can simulate future environmental dynamics and plan over extended horizons (Fig. \ref{fig:game1}). 
For example, inspired by the DreamerV3 agent in Minecraft \cite{hafner2023mastering}, which plans multi-step resource collection and crafting sequences, a UAV could imagine that ``serve user A now, then swing eastward to reach user B, whose video stream will surge in 5 minutes, then ascend to avoid a newly predicted blockage." This imagination-based planning can lead to globally efficient strategies that outperform greedy heuristics or myopic learning agents.
Similarly, the authors in \cite{ho2024uav} utilized predicted user locations to assist communication to the ground users who are unable to get coverage from the base station, reducing the probability of users having no coverage by between 45\% and 85\% compared to non-predictive approaches.

\subsubsection{RIS-Assisted Trajectory and Phase Optimization}

Reconfigurable intelligent surfaces (RIS) introduce new degrees of freedom to LAWNs by dynamically controlling signal reflection to enhance UAV-ground communication. In RIS-assisted UAV systems, the core challenge lies in jointly optimizing the UAV’s 3D trajectory and the RIS phase shifts to maximize metrics such as throughput, energy efficiency, or physical-layer security \cite{zhao2024path}. This forms a high-dimensional and tightly coupled decision problem: the UAV's position affects the angle-of-arrival at the RIS \cite{qin2023joint}.

A world model can capture the nonlinear mappings from UAV-RIS configurations to channel quality metrics without requiring analytical models \cite{assran2025v}. This latent transition model supports temporal prediction of dynamics, enabling the agent to simulate several steps into the future and uncover long-term synergies \cite{ha2018world}. Similar to CarRacing agents trained with Dreamer learn to anticipate and adjust for upcoming curves based on latent state predictions \cite{hafner2019dream}, a UAV-RIS agent can forecast how phase shifts and UAV maneuvers will co-evolve to reinforce signal strength.
Moreover, world models support generative data augmentation by synthesizing diverse virtual channel environments, such as different user layouts or terrain-induced blockage scenarios~\cite{macaluso2024small}. This broadens the training distribution and enhances policy generalization. This is analogous to how Minecraft agents explore procedurally generated worlds with variable landscapes, enabling robust learning across highly dynamic and spatially diverse tasks~\cite{hafner2023mastering}.
Similarly, the authors in \cite{eskandari2024gans} introduced UAV-GAN to augment trajectory planning data, demonstrating how synthetic environments can improve UAV path optimization.

\subsection{Internet of Things}

\subsubsection{Edge Caching}

Edge caching has emerged as a key solution in IoT-enabled networks to reduce latency and backhaul traffic by pre-storing popular content at the network edge. Applications such as adaptive video streaming and firmware delivery benefit significantly from caching strategies that minimize delivery delay and energy consumption while respecting storage and computing constraints \cite{liu2023qoe}. 
Recent works have introduced multi-agent deep reinforcement learning (MADRL) frameworks, such as MacoCache \cite{wang2020intelligent}, to learn cooperative caching policies. 

World models can enhance these intelligent caching frameworks by enabling agents to simulate future demand trajectories \cite{bar2025navigation}. 
For example, edge nodes equipped with a world model can imagine that ``Content X will become popular in geographical area Y within the next hour" and cache accordingly, thereby avoiding the latency and cost of reactive placement.
In addition to temporal prediction, world models enable long-horizon evaluation of caching policies, simulating future user requests to estimate cumulative cache hits. This is akin to how model-based agents in Atari games simulate long action sequences to evaluate rewards under delayed feedback \cite{hafner2020mastering}. Similarly, SwiftCache \cite{abolhassani2024swiftcache} applies model-based learning to dynamic content caching, achieving near-optimal cost under Poisson arrivals and strong performance with limited cache sizes.

\subsubsection{Multi-Hop Routing and Distributed Task Forwarding}

Multi-hop routing is essential in IoT networks where edge nodes or sensors must relay data or computation tasks over multiple intermediate devices to reach a destination with processing or storage capabilities \cite{deng2023multi}. 
The optimization challenge lies in selecting routing paths that minimize end-to-end latency or energy consumption while coping with dynamic topology, partial observability, and shared channel interference~\cite{kong2022edge}.

World models can improve learning-based routing strategies by enabling agents to simulate the future evolution of network states \cite{zhao2025world}. 
Since decisions at one hop influence downstream delay and congestion, an agent equipped with a predictive model can simulate the entire packet trajectory before making a decision. This is similar to how Minecraft agents plan multi-step tool-building sequences to achieve a long-term reward \cite{hafner2023mastering}. Such holistic rollouts allow routing agents to optimize globally rather than myopically.
The authors in~\cite{dhurandher2018location}
considers the node’s present location and the corresponding source to predict the node’s next location or region.
The prediction results determine the probability of a node moving towards the destination, improving routing performance over traditional methods~\cite{dhurandher2018location}.

\subsection{Network Functions Virtualization}

\subsubsection{URLLC/eMBB Coexistence Scheduling}

In network functions virtualization (NFV), 5G-enabled industrial wireless networks must simultaneously serve two heterogeneous traffic classes: ultra-reliable low-latency communication (URLLC) for real-time control and enhanced mobile broadband (eMBB) for high-throughput tasks such as video monitoring \cite{khan2022urllc}. The 3GPP standards permit puncturing, wherein URLLC packets may preempt eMBB transmissions at the mini-slot level, thus ensuring URLLC deadlines at the cost of potential eMBB performance degradation. This scheduling challenge has been studied under various models, including multiple-input multiple-output (MIMO) non-orthogonal multiple access (NOMA) resource allocation, spatial preemptive scheduling, and convex decompositions of user and power assignments~\cite{chen2022coexistence}.
Recent methods incorporate dynamic traffic-aware slicing or Lyapunov-based drift-plus-penalty optimization to balance URLLC latency against eMBB throughput~\cite{feng2020dynamic}. 


World models can enhance coexistence scheduling by learning temporal patterns of URLLC arrivals and their effect on future system states. A world model can forecast traffic bursts that correlate with production cycles, allowing the scheduler to reserve mini-slot headroom in advance \cite{chen2022coexistence}. This long-horizon planning resembles how Dreamer agents in Minecraft simulate multi-step resource gathering paths by modeling future world states \cite{hafner2023mastering}. Analogously, a scheduler equipped with a world model might anticipate that ``a robot arm will trigger URLLC flows every 100 ms” and proactively smooth eMBB allocations beforehand.
Moreover, uncertainty-aware imagination is essential under stochastic URLLC arrivals. Instead of relying on a single-point forecast, the scheduler can simulate diverse future traffic sequences and evaluate how different slicing strategies perform across them \cite{zhang2020stochastic}. This mirrors planning in Atari games such as Montezuma's Revenge, where agents must explore multi-step action paths under unknown risks, balancing exploration and safety \cite{hafner2020mastering}. 


\subsubsection{Elastic Network Slicing}

5G/6G slicing in NFV enables multiple service types, such as URLLC, eMBB, and massive machine type communications (mMTC), to coexist through logically isolated resource partitions. Elastic network slicing refers to dynamically adjusting resource shares, such as bandwidth, computing units, across slices based on real-time demands, maximizing network utilization while meeting slice-specific QoS constraints \cite{gharehgoli2023ai, saibharath2023joint}.

World models can enhance slicing agents by modeling the latent dynamics of traffic patterns and resource contention. For instance, a world model may forecast that ``during the next shift change, mMTC load will spike while eMBB demand drops,” enabling anticipatory slice reallocation. This is analogous to agents in Minecraft who reconfigure tool usage and exploration strategy before entering resource-scarce biomes~\cite{hafner2023mastering}. In this context, world-model-based agents can optimize long-horizon utility while avoiding transient QoS violations.
Additionally, world models can synthesize rare demand scenarios, such as overlapping URLLC and eMBB surges caused by sensor malfunctions, allowing the slicing policy to train on edge cases that may not appear frequently in real-world data \cite{vafa2024evaluating}. 



\section{Lesson Learned}



\subsection{Comparison with Other Models}

\subsubsection{World Models vs. Digital Twins}


Digital twins are high-fidelity virtual counterparts of physical assets that combine live sensor streams with physics-based models to enable real-time monitoring, performance optimization, and, crucially, predictive maintenance~\cite{barricelli2019survey,semeraro2021digital,zhu2025survey,tang2022survey}.
In contrast, a world model is a learned simulator that enables agents to predict environment dynamics and plan actions. It approximates environment transition dynamics so that an agent can imagine future trajectories and evaluate alternative action sequences over long horizons.
Whereas a digital twin seeks centimetre- and millisecond-level correspondence with its physical counterpart to deliver highly accurate, externally verifiable predictions~\cite{cheng2024toward, shen2025revolutionizing, hu2023adaptive}, a world model shifts modeling into a latent space, prioritizing generalization and computational efficiency. This abstraction equips agents in robotics and autonomous systems with internal foresight for decision-making under uncertainty.

\begin{table*}[htp] \scriptsize
  \centering
  \caption{Comparison of world models with digital twins, the metaverse, and foundation models}
  \label{tab:three_compare}
    \begin{tabular}{m{0.09\textwidth}<{\centering}||>{\raggedright\arraybackslash}m{0.15\textwidth}|m{0.10\textwidth}<{\centering}|m{0.27\textwidth}<{\centering}|m{0.12\textwidth}<{\centering}|m{0.12\textwidth}<{\centering}}
      \hline
      \textbf{Model Type}  &  \textbf{Purpose} & \textbf{Model Scope} & \textbf{Algorithms} & \textbf{Applications} & \textbf{Analogy} \\
       \hline
      \multirow{1}{0.09\textwidth}[0pt]{\centering World Model} &
      Learned simulator for agents’ prediction, planning, and imagination in dynamical environments & Agent-centric model
of a dynamical environment & \begin{itemize}[leftmargin=*]
      \item[\textcolor{blue}{\ding{108}}] World Models \cite{ha2018world}: prediction for game playing
          \item[\textcolor{blue}{\ding{108}}] PlaNet \cite{hafner2019learning}: latent-space planning from pixels
          \item[\textcolor{blue}{\ding{108}}] MuZero \cite{schrittwieser2020mastering}: rule-free tree search for games
      \item[\textcolor{blue}{\ding{108}}] DreamerV3 \cite{hafner2023mastering}: imagined RL across 150+ tasks
      \vspace{-1.0em}
      \end{itemize}& \begin{itemize}[leftmargin=*]
      \item[\textcolor{orange}{\ding{108}}] Robotics control
          \item[\textcolor{orange}{\ding{108}}] Autonomous driving
          \item[\textcolor{orange}{\ding{108}}] UAV navigation
      \vspace{-1.0em}
      \end{itemize} & Agent’s ``mental model”\\
      \hline
        \multirow{1}{0.09\textwidth}[0pt]{\centering Digital Twin} &
      Faithful real-world replica of a physical object, person, or process kept up-to-date with real-world data
      & One specific physical asset or process & \begin{itemize}[leftmargin=*]
      \item[\textcolor{blue}{\ding{108}}] User-Centric Resource Management:  \cite{huang2024digital}: multicast short-video QoE optimization
          \item[\textcolor{blue}{\ding{108}}] Data-Oriented Framework \cite{zhou2025user}: immersive communications in 6G
          \item[\textcolor{blue}{\ding{108}}] Resource Allocation \cite{li2024digital}: collaborative sensing \& industrial IoT 
      \item[\textcolor{blue}{\ding{108}}] Map Management 
      \cite{zhou2024digital}: high-fidelity modeling \& augmented reality
      \vspace{-1.0em}
      \end{itemize}& \begin{itemize}[leftmargin=*]
      \item[\textcolor{orange}{\ding{108}}] Predictive maintenance
          \item[\textcolor{orange}{\ding{108}}] Smart-city infrastructure
          \item[\textcolor{orange}{\ding{108}}] Medical device diagnostics
      \vspace{-1.0em}
      \end{itemize} & “Live mirror” of a machine or process\\
      \hline
      \multirow{1}{0.09\textwidth}[0pt]{\centering Foundation Model} &
      Large-scale models trained on broad data to serve a general knowledge engine usable across various tasks & Domain-agnostic
representation learner with billions of parameters & \begin{itemize}[leftmargin=*]
      \item[\textcolor{blue}{\ding{108}}] GPT-4 \cite{achiam2023gpt}: multimodal reasoning LLM
          \item[\textcolor{blue}{\ding{108}}] PaLM 2 \cite{anil2023palm}: multilingual \& code-friendly LLM
          \item[\textcolor{blue}{\ding{108}}] LLaMA 3 \cite{grattafiori2024llama}: open-source 8B/70B/405B family
      \item[\textcolor{blue}{\ding{108}}] Stable Diffusion 3 \cite{esser2024scaling}: text-to-image diffusion model 
      \vspace{-1.0em}
      \end{itemize}& \begin{itemize}[leftmargin=*]
      \item[\textcolor{orange}{\ding{108}}] Text/image generation
          \item[\textcolor{orange}{\ding{108}}] Code completion
          \item[\textcolor{orange}{\ding{108}}] Multimodal reasoning
      \vspace{-1.0em}
      \end{itemize} & ``Universal scholar” \\
      \hline
    \end{tabular}
\end{table*}

\subsubsection{World Models vs. Foundation Models}

Foundation models are large, general-purpose models trained on broad data for diverse tasks, excelling at language, vision, and multimodal understanding \cite{jiang2025comprehensive, guo2025survey}. In contrast, world models are compact, task-specific simulators that predict environment dynamics through state transitions. While foundation models lack an explicit notion of environment state, world models are built around it. Foundation models often run in the cloud for high-level reasoning, whereas world models can operate on edge devices for real-time decision-making \cite{awais2025foundation, zhou2024wall}. 
For example, emerging world foundation models integrate the broad reasoning capabilities of foundation models with the environment-grounded prediction of world models, enabling agents to leverage both general knowledge and task-specific dynamics~\cite{agarwal2025cosmos}.


In summary, digital twins, foundation models, and world models differ in focus. Digital twins emphasize high-fidelity accuracy for real-time monitoring and maintenance. Foundation models leverage broad knowledge for cross-domain reasoning. World models extract temporal dynamics in latent space for predictive decision-making. Despite these differences, they can complement each other as digital twins can integrate foundation reasoning, world models can draw on foundation knowledge, and hybrid approaches such as world foundation models combine accuracy with adaptive intelligence.
For a more intuitive understanding, we present a comprehensive comparison in the Table \ref{tab:three_compare}.

\begin{table*}[htp] \scriptsize
  \centering
  \caption{Representative Projects of World Models: Scenarios and Public Implementations}
  \label{tab:code}
    \begin{tabular}{
    >{\centering\arraybackslash}m{0.09\textwidth} |
    >{\centering\arraybackslash}m{0.10\textwidth} |
    >{\raggedright\arraybackslash}m{0.23\textwidth} |
    >{\centering\arraybackslash}m{0.09\textwidth} |
    >{\centering\arraybackslash}m{0.10\textwidth} |
    >{\raggedright\arraybackslash}m{0.23\textwidth}}
      \hline
      \textbf{Project Name}  &  \textbf{Scenario} & \textbf{Link} & \textbf{Project Name}  &  \textbf{Function} & \textbf{Link} \\
       \hline
    World Model \cite{ha2018world} & Video games & worldmodels.github.io/& PlaNet \cite{hafner2019learning} & Control robotics & github.com/google-research/planet \\\hline
MuZero \cite{schrittwieser2020mastering} & Video games & github.com/werner-duvaud/muzero-general & Dreamer \cite{hafner2019dream} & Control and video games & github.com/google-research/dreamer \\\hline
DreamerV2 \cite{hafner2020mastering} & Video games & github.com/danijar/dreamerv2 & DreamerPro \cite{deng2022dreamerpro} & Continuous control & github.com/fdeng18/dreamer-pro \\\hline
DreamerV3 \cite{hafner2023mastering} & Video games & github.com/danijar/dreamerv3 & TransDreamer \cite{chen2022transdreamer} & Video games &github.com/changchencc/TransDreamer \\\hline
SafeDreamer \cite{huang2023safedreamer} & Continuous control & github.com/PKU-Alignment/SafeDreamer &  SGF \cite{robine2025simple} & Video games & github.com/jrobine/sgf\\\hline
V-JEPA \cite{bardes2023v} & Video understanding & github.com/facebookresearch/jepa & 
    V-JEPA 2~\cite{assran2025v}& Control robotics & github.com/facebookresearch/vjepa2 \\
    \hline
    \end{tabular}
\end{table*}


\subsection{World Model for Edge General Intelligence}

EGI requires agents that can \emph{reason, predict, and act} reliably under the tight latency, energy, and privacy constraints of the edge. The heart of such autonomy could lie in the world models. When integrated inside the cognition module of an agentic AI loop, world models empower edge agents with internal physics engines, allowing them to simulate and optimize long-horizon strategies on-device.

\subsubsection{A cognitive backbone for agentic AI} 

By encoding raw multimodal observations into latent states, world models enable a closed loop of perception, prediction, and control. These latent dynamics facilitate hierarchical reasoning, where low-level actions are refined internally and high-level plans from agentic AI are validated for feasibility locally. For instance, a UAV agent equipped with a world model can anticipate signal fading, traffic surges, or weather shifts without cloud queries, mirroring the foresight seen in CarRacing or Dreamer agents~\cite{ha2018world, hafner2019learning}.

\subsubsection{Synergy with foundation models and digital‑twin models} Whereas foundation models supply declarative knowledge and digital twins deliver high‑fidelity global context, the world model forms an action‑conditioned bridge between the two. For instance, in autonomous driving, a cloud twin furnishes city‑scale traffic priors. A local world model, such as NIO WorldModel, forecasts sub‑second collision risks, and an LLM planner narrates route‑level intents—together realising real‑time cognition without saturating the link~\cite{zhao2025generative, hu2023gaia}.

\subsubsection{Key enablers for EGI deployment}

World models unlock several critical capabilities essential for EGI:
\begin{itemize}
    \item \textbf{On-Device Autonomy:} Unlike LLMs with billion-scale parameters, latent dynamics models preserve only task-relevant variables, allowing real-time execution on edge devices like UAVs or industrial controllers \cite{hafner2019dream}.
    \item \textbf{Predictive Scheduling:}
    Agents equipped with world models can simulate hundreds of imagined futures in latent space, enabling multi-minute planning with only sparse cloud synchronization \cite{hafner2019dream}.
    \item \textbf{Counterfactual Resilience:}
    By learning generalized transition rules rather than memorizing samples, world models synthesize novel, hypothetical states, vital for robustness in unanticipated edge scenarios \cite{zhao2025world}.
\end{itemize}

In summary, world models transform edge devices into self-improving cognitive systems, bridging perception and foresight within a single, latency-aware architecture. They represent a paradigm shift from reactive inference to proactive intelligence for EGI.
For convenience, we also provide a summary of the relevant open-source code in Table \ref{tab:code}.



\subsection{Research Challenges And Future Directions}

While world models offer immense promise for Edge General Intelligence, several technical barriers hinder their widespread deployment, particularly in edge settings where resources are constrained, latency is critical, and environments are highly dynamic.

\subsubsection{Compute and Memory Bottlenecks}

Current video-based world models often require massive computation and storage, with parameter counts reaching billions and inference latencies measured in seconds \cite{bar2025navigation, russell2025gaia}. These models remain impractical for edge platforms lacking GPU/TPU acceleration. For example, the NWM diffusion model needs desktop-class GPUs to sustain real-time planning, which exceeds typical UAV or sensor hardware capabilities.


To address this, researchers have explored several approaches, including \textbf{quantization} for low-bit inference~\cite{lee2023cqm}, \textbf{knowledge distillation} into lightweight models~\cite{yamada2024twist}, and \textbf{architecture innovations} such as CDiT, which reduces complexity by four times without compromising fidelity~\cite{bar2025navigation}.

\subsubsection{Real-Time Planning Under Constraints}


Many edge applications, such as autonomous driving or mmWave beam alignment, require sub-100-ms decisions, yet pixel-based world models often fall short of meeting this real-time constraint. To overcome this limitation, hybrid strategies have been proposed, including \textbf{asynchronous background simulation} to precompute imagined rollouts~\cite{zhang2019asynchronous}, \textbf{heuristic-lightweight frontends} that trigger full model inference only under complex conditions~\cite{gumbsch2023learning}, and \textbf{trajectory- or vector-level abstraction}, which accelerates planning while preserving essential dynamics~\cite{zhang2023trafficbots}.

\subsubsection{Robustness in Out-of-Distribution Environments}

Another critical issue is that world models trained on narrow or biased datasets may hallucinate when encountering novel conditions, such as rare weather or unseen mobility patterns. For instance, a recent study shows that even large video diffusion models revert to case-based imitation under out-of-distribution scenarios~\cite{kang2024far}. Mitigation strategies include employing diverse and synthetic training datasets to capture edge cases without risking real systems, enabling online adaptation through few-shot or continual learning with methods such as LoRA or adapter modules~\cite{chen2024superlora}, and adopting modular fine-tuning techniques that allow local adjustments without retraining the entire model~\cite{yu2024edge}.

\section{Conclusion}



World models offer a powerful paradigm for bridging perception and foresight in edge environments, enabling agentic AI systems to act with strategic autonomy rather than reactive heuristics. By learning compact representations of environmental dynamics and supporting imagination-driven planning, world models transform edge devices into proactive decision-makers capable of anticipating future states, reasoning under uncertainty, and optimizing long-term objectives. This survey has provided a unified framework that connects world model theory with EGI applications, identifying core architectures, planning methods, and real-world deployments across vehicular networks, LAWNs, IoT networks, and beyond. 
While recent advances demonstrate promising directions, significant challenges remain in making world models lightweight, safe, and generalizable for constrained edge deployments. 
This survey concludes by summarizing promising research directions and unresolved challenges in applying world models to EGI, aiming to guide future efforts toward practical and scalable deployments.

\bibliography{Ref}

\end{document}